\documentclass[10pt,journal,compsoc]{IEEEtran}
\ifCLASSOPTIONcompsoc
  \usepackage[nocompress]{cite}
\else
  \usepackage{cite}
\fi
\usepackage{xspace}
\usepackage[normalem]{ulem}
\usepackage{verbatim}

\usepackage{booktabs}
\usepackage{multirow}

\usepackage{fontawesome}
\usepackage{algorithm}  
\usepackage{algorithmic} 
\usepackage{graphicx}
\usepackage{url}
\usepackage{hyperref}

\usepackage[dvipsnames]{xcolor}
\definecolor{fast}{HTML}{ED6940}
\definecolor{slow}{HTML}{FF903E}
\definecolor{louder}{HTML}{84B13D}
\definecolor{softer}{HTML}{B7CD49}
\definecolor{pause}{HTML}{4387ED}
\definecolor{stress}{HTML}{9A86D6}
\definecolor{disconnection}{HTML}{feb24c}
\definecolor{repetition}{HTML}{3182bd}
\definecolor{polarity}{HTML}{538d22}
\definecolor{subjectivity}{HTML}{3a6ea5}
\definecolor{alliterations}{HTML}{720026}
\definecolor{rhymes}{HTML}{CE4257}

\newcommand{\louder}{\textcolor{louder}{\faLevelUp}}

\newcommand{\pause}{\textcolor{pause}{\faSquare}}
\newcommand{\stress}{\textcolor{stress}{\faUnderline}}
\newcommand{\disconnection}{\textcolor{disconnection}{\faCircleONotch}}
\newcommand{\repetition}{\textcolor{repetition}{\faRefresh}}
\newcommand{\polarity}{\textcolor{polarity}{\faArrowsV}}
\newcommand{\subjectivity}{\textcolor{subjectivity}{\textbf{( )}}}

\usepackage[running]{lineno}

\newcommand{\fontsmall}{\fontsize{8pt}{10pt}\selectfont}

\newcommand{\qbox}[1]{%
	\medskip
	\fcolorbox{gray}{white}{
		\begin{minipage}{0.91\linewidth}
			\begin{internallinenumbers}
			\resetlinenumber
			\fontsmall
			\emph{#1}
			\end{internallinenumbers}
		\end{minipage}
	}
	\medskip
}

\newcommand{\reducedstrut}{\vrule width 0pt height .9\ht\strutbox depth .9\dp\strutbox\relax}
\newcommand{\colbox}[2]{
\begingroup
  \setlength{\fboxsep}{0pt}%
  \colorbox{#1}{\reducedstrut#2\/}%
  \endgroup}
  
\definecolor{gblue}{rgb}{0.12, 0.47, 0.70}
\definecolor{gorange}{rgb}{1, 0.5, 0.05}
\definecolor{ggreen}{rgb}{0.17, 0.63, 0.17}
\definecolor{ggrey}{rgb}{0.5, 0.5, 0.5}
\definecolor{gpurple}{rgb}{0.58, 0.4, 0.74}
\definecolor{bleudefrance}{rgb}{0.19, 0.55, 0.91}

\usepackage[most]{tcolorbox}
\newtcbox{\inlinebox}[1][]{enhanced,
 box align=base,
 nobeforeafter,
 colback=white,
 opacityback=1.0,
 colframe=bleudefrance,
 size=fbox,
 left=1pt,
 right=1pt,
 boxsep=2pt,
 #1}

\newcommand{\casetag}[1]{\inlinebox{\textcolor{bleudefrance}{\fontfamily{lmss}\selectfont{#1}}}}

\newcommand{\gitaly}[1]{\textcolor{white}{\colbox{ggrey}{#1}}}
\newcommand{\ggermany}[1]{\textcolor{white}{\colbox{ggreen}{#1}}}
\newcommand{\gstigma}[1]{\textcolor{white}{\colbox{gblue}{#1}}}
\newcommand{\gevil}[1]{\textcolor{white}{\colbox{gorange}{#1}}}

\newcommand{\example}[1]{\emph{``{#1}''}}

\usepackage{paralist}
\newcommand{\name}{{\textit{DeHumor}}}
\newcommand{\cp}{\textit{{control panel}}}
\newcommand{\hx}{{\textit{humor exploration}}}
\newcommand{\Hx}{{\textit{Humor exploration}}}
\newcommand{\hf}{{\textit{humor focus}}}

\newcommand{\ie}{i.e.}
\newcommand{\eg}{e.g.}
\newcommand{\etal}{et al.}

\newcommand{\xingbo}[1]{{\color{black} #1}}
\newcommand{\xb}[1]{{\color{black} #1}}
\newcommand{\xbRevise}[1]{{\color{black} #1}}
\newcommand{\xbRev}[1]{{\color{black} #1}}

\usepackage{enumitem}
\setitemize{noitemsep,topsep=0pt,parsep=0pt,partopsep=0pt}
\usepackage{fancyhdr}

\usepackage{hyperref}

\usepackage{lipsum}
\usepackage{color}

\usepackage{tikz}

\newcommand\copyrighttext{%
  \footnotesize \textcopyright 2021 IEEE. Personal use of this material is permitted.
  Permission from IEEE must be obtained for all other uses, in any current or future
  media, including reprinting/republishing this material for advertising or promotional
  purposes, creating new collective works, for resale or redistribution to servers or
  lists, or reuse of any copyrighted component of this work in other works.
  DOI: \href{10.1109/TVCG.2021.3097709}{10.1109/TVCG.2021.3097709}}
\newcommand\copyrightnotice{%
\begin{tikzpicture}[remember picture,overlay]
\node[anchor=south,yshift=10pt] at (current page.south) {\fbox{\parbox{\dimexpr\textwidth-\fboxsep-\fboxrule\relax}{\copyrighttext}}};
\end{tikzpicture}%
}

\begin{document}

%
\title{DeHumor: Visual Analytics for\\ Decomposing Humor}
%
%
%
%

\author{Xingbo~Wang, 
        Yao~Ming, 
        Tongshuang~Wu, 
        Haipeng~Zeng, 
        Yong~Wang, 
        and~Huamin~Qu
\IEEEcompsocitemizethanks{\IEEEcompsocthanksitem X. Wang and H. Qu are with the Hong Kong University of Science and Technology.\protect\\
E-mail: \{xwangeg,~huamin\}@cse.ust.hk
\IEEEcompsocthanksitem Y. Ming is with Bloomberg LP.\protect\\
E-mail: yming7@bloomberg.net
\IEEEcompsocthanksitem T. Wu is with the University of Washington.\protect\\
E-mail: wtshuang@cs.washington.edu
\IEEEcompsocthanksitem H. Zeng is with the Sun Yat-sen University.\protect\\
E-mail: zenghp5@mail.sysu.edu.cn
\IEEEcompsocthanksitem Y. Wang is with the Singapore Management University.\protect\\
E-mail: yongwang@smu.edu.sg
}
\thanks{Manuscript received September 28, 2020; revised July 18, 2021.}}

%
%

\markboth{IEEE Transactions on Visualization and Computer Graphics}%
{Shell \MakeLowercase{\textit{et al.}}: Bare Demo of IEEEtran.cls for Computer Society Journals}
%



\IEEEtitleabstractindextext{%
\begin{abstract}
Despite being a critical communication skill, grasping humor is challenging---a successful use of humor requires a mixture of both engaging content build-up and an appropriate vocal delivery (\eg, pause).
Prior studies on computational humor emphasize the textual and audio features immediately next to the punchline, yet overlooking longer-term context setup.
Moreover, the theories are usually too abstract for understanding each concrete humor snippet.
To fill in the gap, we develop \name, a visual analytical system for analyzing humorous behaviors in \xbRev{public speaking}.
To intuitively reveal the building blocks of each concrete example, {\name} decomposes each humorous video into \xbRev{multimodal} features and provides inline annotations of them on the video script.
In particular, to better capture the build-ups, we introduce content repetition as a complement to features introduced in theories of computational humor and visualize them in a context linking graph.
To help users locate the punchlines that have the desired features to learn, we summarize the content (with keywords) and humor feature statistics on an augmented time matrix.
With case studies on stand-up comedy shows and TED talks, we show that {\name} is able to highlight various building blocks of humor examples.
In addition, expert interviews with communication coaches and humor researchers demonstrate the effectiveness of {\name} for \xbRev{multimodal} humor analysis of speech content and vocal delivery.
\end{abstract}

\begin{IEEEkeywords}
Humor, Context, \xbRev{Multimodal Features}, Visualization.
\end{IEEEkeywords}}

\maketitle
\copyrightnotice

\IEEEdisplaynontitleabstractindextext

%
\IEEEpeerreviewmaketitle

\IEEEraisesectionheading{\section{Introduction}\label{sec:introduction}}
\IEEEPARstart{H}{umor}---the use of puns, turns of phrases, or humorous anecdotes---is a powerful communication skill for public speakers to connect, engage, and entertain their audiences.
A proper usage of humor can help induce shared amusement~\cite{mulholland2003handbook}, reduce social anxiety~\cite{wooten1996humor}, and boost persuasive power~\cite{davidson2003complete}.
Although \emph{identifying} humor (\ie, judging whether a joke is funny or not) comes natural to us, \emph{becoming} humorous is challenging in practice, as it requires the integration of various humor skills. 
To create humorous content (\ie, jokes), the speakers need to come up with intelligent gradual setups, as well as a sudden twist to subvert the audience's expectation~\cite{raskin2012semantic}.
Then, to effectively achieve the intended dramatic effect, speakers have to decorate the contents with appropriate acoustic delivery methods~\cite{jefferson1979technique,bauman1986story}
(\eg, pause, pitch).
Thorough understanding and learning of humor can only be achieved if we can decompose these building blocks and the interactions between them.

Prior work across multiple disciplines (psychology, philosophy, and linguistics) has qualitatively \xbRev{characterized} humor.
For example, 
Plato described humor as an expression of superiority to others,
while
Schopenhauer~\cite{schopenhauer1891world} stated that humor comes from the realization of incongruous interpretations of a statement.
These abstract theories become the cornerstone of various computational features that capture phonetic~\cite{mihalcea2005making,yang2015humor},
stylistic~\cite{mihalcea2005making,mihalcea2006learning}, human-centered~\cite{yang2015humor, mihalcea2007characterizing,zhang2014recognizing}, and content-based~\cite{taylor2009computational,radev2015humor} humor characteristics.

While these features make it possible to quantify humor, analyzing concrete humor examples \xbRev{in speeches} remains challenging for two reasons.
First, presenting a laundry list of features \xbRev{for a humorous speech} can be overwhelming.
Not only do the features create a heavy perceptual burden, but the sophisticated interactions between different building blocks remain hidden in the large feature space.
Most of the analyses to date~\cite{mihalcea2005making, mihalcea2007characterizing,yang2015humor, pickering2009prosodic,attardo2011prosodic,attardo2011timing,attardo2013multimodality}
only focus on either humorous content or vocal delivery independently. However,
\xbRevise{both of them are needed simultaneously to understand humor in practice.}
A \xingbo{\textbf{punchline}---the most important sentence that triggers the audience response (\eg, laughter)---}may be mundane in isolation but hilarious when rendered with exaggerated volume and pitch.
In the example\footnote{The example link: \xingbo{\url{http://bit.ly/2MrVKf9}}} below,
patterns like acoustic stressing at the modal particles (\ie, getting louder at \example{okay} in \xbRev{Line \#5}) or pausing to emphasize turning points (\ie, pausing before \example{I} in Line \#4 to contrast with \example{they} in the preceding sentence) are only observable when we highlight their occurrences.


\qbox{So when I show up to a crime scene, \\
Somebody is always like, ``are you a cop?''\\
I don't wanna say I'm a cop cause it's against the law.\\
So they go, ``are you a cop?''\\
And I go, \textcolor{gray}{\texttt{[PAUSE]}} ``I'll ask the f**king questions, okay\textcolor{gray}{\texttt{[LOUDER]}}?''}

\vspace{-2mm}

Second, the already overwhelming feature definitions have yet to be comprehensive. 
Existing research \xbRev{emphasizes} short jokes (\eg, one-liners) while overlooking those with longer-term set-ups, making it difficult to track the clues that help lead to the core punchline.
In the previous example, 
the punchline in Line \#5 only becomes funny after Lines \#1 to \#5 \xbRev{provide} the essential context:
When attempting to enter a crime scene (\#1), the speaker was asked about the cop identity (\#2). 
Not being an actual cop, he avoided explicitly making an illegal claim that he is one (\#3).
As a workaround, he quoted a common trope for police in movies and TV (\emph{I'm asking the f**king question!}, \#5), and so to mislead the \example{somebody} to believe that he is a cop.
This example demonstrates the importance of contexts for humor analysis, which motivates our study.

To better decompose humor examples into critical features, as well as intuitively present the mixtures of these features, we present a novel visual analytics system named {\name}.
{\name} \xingbo{aims to help domain experts (\eg, communication coaches and researchers) analyze the verbal content and vocal delivery of}
\xbRev{public speaking} containing many humorous punchlines (labeled with audience response markers like \texttt{[LAUGHTER]}).
We formulate the design requirements based on literature surveys and user interviews with five humor researchers as well as two communication coaches. We choose to interview these experts because they have theoretical knowledge in humor and need assistance on a systematic investigation of humor.
Accordingly, we design {\name} to support \xingbo{multi-level explorations of humorous texts and delivery in speeches.}
We aim to enable users to easily understand when and where humorous punchlines are inserted (speech-level), how one particular punchline relates to its preceding build-up sentences (context-level), and how the \xbRev{vocal delivery}
and the textual content are paired within each sentence (sentence-level).
In particular, to reveal the interactions between textual and audio features, we provide inline annotations of the features along with the raw \xingbo{transcripts}.
To highlight the build-ups, we introduce a context linking graph that can recognize relevant phrases and visually connect them \xingbo{with links}.
With case studies on stand-up comedy shows and TED talks, we show that {\name} can highlight various building blocks of humor examples.
Interviews with domain experts further confirm that {\name} is helpful for exploratory and in-depth analysis of humorous snippets.

In summary, the major contributions of our work are:
\begin{compactitem}
    \item We design a visual analytics system to support interactive and multimodal analysis of humorous pieces and reveal humor strategies in speech content and voice.
    \item We demonstrate the usability and effectiveness of {\name} through case studies and expert interviews with communication coaches and humor researchers.
    
\end{compactitem}
\section{Related Work}
This section reviews related research on humor theory and visualization for speech analysis.
\subsection{Computational Humor Features on Speech}
\label{subsec:relate_comp_humor}
Modeling humor features is beneficial and critical for automatic humor understanding. Prior work has modeled humor using both text and audio features.
For textual features, 
Mihalcea and Strapparava~\cite{mihalcea2005making}
extracted stylistic features that characterize humorous texts, including alliteration, antonym, and adult slang. 
Later, Mihalcea and Pulman~\cite{mihalcea2007characterizing}
extended feature sets with human centeredness and polarity orientation.
Kiddon and Brun~\cite{kiddon2011s} 
measured erotica-level of nouns, adjectives, and verb phrases.
Zhang and Liu~\cite{zhang2014recognizing}
designed five categories of humor-related linguistic features, including morpho-syntactic features, lexico-semantic features, pragmatic features, and phonetic features\cite{yang2015humor,donahue-etal-2017-humorhawk}.
Content-based features (e.g., n-grams~\cite{taylor2004computationally, yan2017duluth}, lexical centrality~\cite{radev2015humor}, incongruity~\cite{yang2015humor}, and word associations~\cite{cattle2018recognizing}) were also widely experimented to study the patterns in humorous text content in previous work.
However, most of these features were not systematically derived and were defined in an empirical way.
Yang \etal~\cite{yang2015humor} proposed a computational framework
to describe the latent semantic structures of humor, including incongruity structure, ambiguity theory, interpersonal effect, and phonetic style.
Bali \etal~\cite{ahuja2018makes} extracted three major characteristics across all humor types, which are mode, theme, and topic.
The mode (\eg, exaggeration) is dependent on situations of delivery. The theme relates to emotions behind the use of language. The topic covers the central elements of humor.
In our work, we integrate and extend the above frameworks for textual features to analyze humorous texts.

For audio features,
previous quantitative studies~\cite{pickering2009prosodic,attardo2011prosodic,attardo2011timing,attardo2013multimodality,purandare2006humor,bertero2016long} identified significant dimensions for joke-telling, 
including volume, pitch, speech rate, and pause length.
Pickering \etal~\cite{pickering2009prosodic} found punchlines were produced with lower pitch in joke narrative.
Attardo and Pickering~\cite{attardo2011timing} investigated pauses around punchlines.
Purandare and Litman~\cite{purandare2006humor} used acoustic-prosodic features (\ie, pitch, energy, and tempo) and linguistic features to automatically recognize the humor in the TV sitcom.
Bertero and Fung~\cite{bertero2016long}
modeled conversational humor by combining
speech utterances with a set of high level features (\eg, speaking rate).
Our work computes pitch, volume, speed, and pauses around punchlines and their contexts, to reveal acoustic patterns in the humor delivery.


\subsection{Visualization for Speech Analysis}
Speech visualization is an important research topic in the multimedia analysis.
It is applied in many domains, such as 
public speaking training~\cite{wang2020voicecoach,rubin2015capture}, 
visualization for the hearing impaired~\cite{watanabe2000speech}, 
and emotion analysis~\cite{zeng2019emoco}.
\xingbo{While 
some prior studies have visualized
the speaker/audience interactions and topic dynamics in multi-party speeches (\eg, debate, conversation)~\cite{rony2020claimviz, el2016contovi, debatevis2020}, our work focuses on analyzing verbal content (\eg, word use) and vocal delivery (\eg, voice modulation) of humor in public speaking.}

One of the main goals of visualizing speech data is to intuitively and effectively reveal the relationship between content and speaking voice.
The most straightforward way is to encode sequential features as bar charts or line charts and then draw them along the script~\cite{zeng2019emoco,
oktem2017prosograph}.
Oh~\cite{oh2010text} used a vertical timeline to summarize features of sections in songs.
However, directly overlaying features on the words can lead to a high cognitive load. Moreover, it does not explicitly demonstrate relationships between words.
Patel and Furr~\cite{patel2011readn} 
\xbRev{proposed 
a method to directly encode the prosodic features using text properties.}
It manipulates the vertical offset, opacity, and letter spacing of texts to represent pitch, intensity and audio duration, respectively. 
Similarly, Wang \etal~\cite{wang2020voicecoach} and Rubin \etal~\cite{rubin2015capture} designed intuitive glyphs to represent prosodic features, which annotates speakers' vocal performance on the script.
Similarly, in our work,
we design glyphs and adjust text styles to explicitly demonstrate the humor features and their relationships.

Besides, we aim to visualize the semantic relationship of texts in humor snippets to help understand textual humor.
Here, we summarize \xbRev{the prior studies that have} inspired our research.
Matrices~\cite{cao2016introduction,janicke2014visualizations} are widely adopted to visualize co-occurrence patterns in text documents.
Word clouds~\cite{vuillemot2009s} can also summarize word relations.
It is also common to use graphs~\cite{wattenberg2002arc,don2007discovering} and links~\cite{subavsic2008web,riehmann2015visual,sinclair1991corpus} to describe co-occurrence and repetitions.
However, it is challenging to directly apply these techniques in our work.
For example,
word clouds lose temporal information.
Matrices suffer from space inefficiencies, especially when they are sparse. 
Arc diagrams alleviate the above issues by placing words in a line and visually connecting them.
Still, if the text is long, it is difficult to obtain an overview of the text relationship while keeping the temporal orders.
In this paper,
we extend the arc diagram with a multi-level context summary and rich interactions to support the effective identification of \xingbo{contextual repetitions} in a humor snippet.




\section{Design Process}
\label{sec:designprocess}
\begin{table*}
\caption{A summary of humor-related features that we identify from the qualitative and quantitative research of humor.}
\label{table:featuretable}
\vspace{-3mm}
\centering
\scalebox{0.8}{
\begin{tabular}{@{}cllll@{}}
\toprule
\multicolumn{1}{r}{} &
  \textbf{Humor-related features} &
  \textbf{Subcategory} &
  \textbf{Description} &
  \textbf{References} \\ \midrule
\multirow{8}{*}{Textual} & Content-related features        &               & Key concepts (e.g. situation) on which the humorous story is built                     & \cite{ahuja2018makes, mihalcea2005making, vuillemot2009s, yang2015humor}  \\ \cmidrule(l){2-5} 
                         & \multirow{2}{*}{Incongruity}    & Disconnection & \xingbo{Semantic disconnection (e.g., contrast) between two content words in a sentence}        &  \cite{mihalcea2005making, yang2015humor,  radev2015humor, yuan2008speaker} \\ \cmidrule(l){3-5} 
 &
   &
  \xingbo{Intra-sentence repetition} &
  Repeating similar objects in a sentence & \cite{mihalcea2005making, radev2015humor, vuillemot2009s, yang2015humor, yuan2008speaker}
   \\ \cmidrule(l){2-5} 
 &
  \multirow{2}{*}{Human-centeredness} &
  Polarity &
  Positive/negative orientation of emotion & \cite{mihalcea2005making, yuan2008speaker, reyes2012humor, zhang2014recognizing, yang2015humor}
   \\ \cmidrule(l){3-5} 
 &
   &
  Subjectivity &
  Subjective/objective orientation & \cite{yuan2008speaker, yang2015humor, davis2008communication, castro2017crowd}
   \\ \cmidrule(l){2-5} 
                         & \multirow{2}{*}{Phonetic style} & Alliteration  & Occurrences of the same letter or sound at the beginning of a group or words & \cite{yang2015humor, donahue-etal-2017-humorhawk, mihalcea2005making, yuan2008speaker, zhang2014recognizing}   \\  \cmidrule(l){3-5} 
 &
   &
  Rhyme &
  Repetition of similar sounds in the final stressed syllables of a group of words & \cite{donahue-etal-2017-humorhawk, yang2015humor, yuan2008speaker, zhang2014recognizing}
   \\ \cmidrule(l){2-5} 
                         & Build-ups                       & \xingbo{Inter-sentence repetition}    & Concepts (e.g., a person) that have been previously told before a punchline  &  \cite{bauman1986story, nash2014language, raskin2012semantic} \\ \midrule
\multirow{4}{*}{Audio} &
  Volume &
  Volume variation &
  Variation in volume: softer and louder & \cite{attardo2011prosodic, bertero2016long, purandare2006humor, schuller2010interspeech}
   \\ \cmidrule(l){2-5} 
 &
  Pitch &
  Stress &
  Vocal stress by raising pitch & \cite{attardo2011prosodic, bertero2016long, purandare2006humor, pickering2009prosodic, schuller2010interspeech}
   \\ \cmidrule(l){2-5} 
 &
  Pause &
   &
  A temporary stop in speech & \cite{attardo2011prosodic, bertero2016long, purandare2006humor, schuller2010interspeech}
   \\ \cmidrule(l){2-5} 
 &
  Speed &
  Speed variation &
  Variation in speech rate: faster and slower & \cite{attardo2011prosodic, bertero2016long, purandare2006humor, schuller2010interspeech}
   \\ \bottomrule
\end{tabular}
}
\vspace{-3mm}
\end{table*}

Our goal is to support an in-depth and systematic investigation into the humor composition and its vocal delivery in \xbRev{public speaking} (\eg, oral presentation). 
Our main target \xbRev{users are people}
who have theoretical knowledge in humor and are motivated to study humorous speech (\eg, humor researchers and communication coaches).
We expect our system to alleviate their mental burden when analyzing unstructured humorous speech (\ie, texts and audio) in an organized and quantitative way.

To build a concrete understanding of humor,
we conducted literature reviews and informal user interviews to identify a set of textual and audio features that are both quantifiable and essential for humor analysis. Specifically, we first summarized features from existing literature and proposed a new computational method for extracting the context-related feature (\ie, \xingbo{inter-sentence} repetition)
to supplement the framework of computational humor. 

Next, based on these feature candidates, we interviewed five humor researchers and two communication coaches who provided professional insights into humor study and helped validate our proposed features regarding their importance and helpfulness for humor analysis.
Meanwhile, during interviews, we inquired about their perspectives on the analytical aspects and challenges within humor analysis.
Based on their feedback, we distilled design requirements, which guided our initial system design.
The researchers (\textit{E1-E5}, including three postgraduates, one PhD graduate, and one university lecturer) study English/applied linguistics, English literature, and L2 learning. Three of them have published research papers on humor. They all have done humor research projects.
The communication coaches (\textit{C1, C2}) 
have five and ten years of communication skills training experience, respectively.
\xbRev{One part of their work is to} train speakers to deliver humorous speeches based on pre-selected humor examples or topics.


\subsection{Literature Review \& User Interviews}
\subsubsection{Humor features}

We borrowed \xingbo{the most common and essential quantifiable features} 
of humor content creation and vocal delivery from the existing work mentioned in Sec.~\ref{subsec:relate_comp_humor}.
For the textual features, we organized and selected our features based on
the frameworks proposed by Yang et al.~\cite{yang2015humor} and Bali et al.~\cite{ahuja2018makes}, such that our features cover both semantic structures (\eg, incongruity and phonetic style) and content understanding (\eg, topic).
Similarly, for the audio features, we collected features and merged the related ones from different studies (\eg, tempo~\cite{schuller2010interspeech} v.s. speech rate~\cite{pickering2009prosodic}, and energy~\cite{schuller2010interspeech} v.s. volume~\cite{pickering2009prosodic}).
As a result, we covered four audio aspects: volume, pitch, pause, and speed.
The full list of features is in Table~\ref{table:featuretable}, and the computations are in Sec.~\ref{subsec:humorFeatureExtraction}.

While these features comprehensively summarize different aspects of one-liners, existing computational research rarely covers context features in humor cases with longer set-ups.
For example, \xingbo{\textbf{(inter-sentence) repetition}} is one essential comedic devices~\cite{goldstein1970repetition}.
Consider a simple example in~\cite{tannen2007talking}:
\example{\textit{A}: Rover (a dog) is being good.
\textit{B}: I know.
\textit{C}: He is being hungry.}
The repetition of the structure ``he is being'' makes the audience expect a similar response to A's.
However, the word ``hungry'' conflicts with the expectation, and the repetition enhances the dramatic effect of the twist.
To seize such patterns in the build-up of a humorous story, we introduce an algorithm to compute context-level ``\textit{repetition}''. The detail of the computation is illustrated in Sec.~\ref{subsubsec:context_analysis}.

\subsubsection{User interviews}
To validate the proposed humor feature sets and discover analytical needs for humor analysis,
we conducted independent interviews with the seven target users (\textit{E1-E5}, \textit{C1, C2}) mentioned earlier.
During interviews, we asked the participants to
(1) describe their general process/methodology of humor analysis,
(2) illustrate what aspects of humor in speeches they care about and how do they rank our proposed features regarding the importance/difficulty for analysis,
\xingbo{(3)} propose desired design requirements (\eg, functions) for a system that facilitates systematic humor analysis.
Their feedback is
reported as follows. 
The design requirements are summarized in Sec.~\ref{subsec:designrequirements}.


\textbf{Whole-to-part analysis.}
\xbRev{According to the participants' feedback,}
they generally analyze the speech \emph{from the whole (\eg, speech topics) to the parts (\eg, \xingbo{word use}).}
They usually first search for humor examples from public speeches, TV shows, and books according to humor topics, genres, and comedians.
Then, they 
focus on
the humorous pieces that can elicit laughter from the audience and investigate 
the patterns of speech content and delivery in humor \xbRev{speeches}.
Specifically, the analysis follows the \textbf{language strata}~\cite{halliday2014introduction, martin1992english}---\textit{the context, semantics/pragmatics, lexemes (words and phrases), and phones}---from coarse to fine.

\begin{table}[]
\caption{\xbRevise{The average importance/difficulty rankings for the analytical aspects 
(A smaller rank value means greater importance/difficulty)}.}
\label{tab:rank_question}
\vspace{-3mm}
\centering
\scalebox{0.85}{
\begin{tabular}{l|lllll}
\hline
                     & \textbf{Word usage} & \textbf{Vocal delivery} & \textbf{Build-ups} & \textbf{Timing} & \textbf{Topics} \\ \hline
\textbf{Importance}  & 1.86    & 2.57         & 3.00      & 3.14    & 4.43  \\ \hline
\textbf{Difficulty} & 2.71        & 2.43             & 1.86    & 4.57     & 3.43     \\ \hline
\end{tabular}}
\vspace{-3mm}
\end{table}

\textbf{Analytical aspects and computational features.}
\xbRevise{As shown in Table~\ref{tab:rank_question}, \emph{word usage} (rank: 1.86) and \emph{vocal delivery} (rank: 2.57) with the \xbRev{highest} importance rankings 
were considered essential for humor analysis.}
\xbRev{The participants} appraised the extraction of incongruous words, affective lexicons (\ie, sentiment and subjectivity), \xbRev{and} phonetic styles (\ie, alliteration and rhyme). 
These features cover their typical analytical interests and provide quantitative and concrete evidence for language patterns of humor in semantics, lexemes, and phones.
\textit{E1} claimed that the incongruous words can reflect the unexpected conflicts and twists in punchlines, which are at the core of an influential humor theory---incongruity.
\textit{E3} added that the negative sentiment lexicons help study the styles of self-deprecating or self-enhancing humor. 
The participants also thought the acoustic features---volume, pitch, pause, speed---can effectively
reflect the vocal characteristics of humor. 
\xb{For example, smart pauses (\eg, comic timing) are effective for building up suspense.}
C1 said that ``I can use them (acoustic features) to tell whether a speaker is good at telling jokes or not.''

Besides, the two coaches attached much importance to the \emph{timing of humor}.
They considered it as a good starting point to learn humor in \xbRev{public speaking}.
\textit{C1} reasoned that 
finding a proper place (\eg, speech opening) to insert humor
may be the easiest thing for ordinary people to learn, which can make a big impact on their speeches. 
\textit{E4} suggested the timing should include the distribution and drift of topics (``\textit{What content it supports} and \textit{how the topics evolve}'').

\xbRevise{In terms of difficulty (Table~\ref{tab:rank_question}), \emph{humor build-ups} was regarded as the most difficult aspect with the \xbRev{top} rank (1.86).}
\xbRev{The participants} thought that
the cognitive load of backtracking and comprehending related concepts (\eg, background, characters, emotion) in humor build-ups can be heavy. The inter-sentence repetition extraction was considered reasonable and helpful. \textit{E4} said, ``\textit{It (the context repetition) connects the dots (of humor).}''
\textit{E3} specified, ``\textit{It is useful for revealing the humor structure and can help summarize the core idea of humor.}''
Still, some context-related humor characteristics proposed by the participants, such as the social background, culture, humor genres (\eg, dark humor), are difficult or unreliable to be quantified. Thus, they are left as future research.

\xb{Besides the humor features for word usage, speech content, and vocal delivery (Sec.~\ref{subsec:relate_comp_humor}),
we enrich the existing computational framework with inter-sentence repetitions for humor context analysis and the timing of humor based on the participants' feedback.}
Their computation and visualization are described in Sec.~\ref{subsubsec:context_analysis} and Sec.~\ref{subsec:augmented_timematrix} respectively.

\textbf{Desired functionality.}
Since there are few tools that 
enable humor exploration and analysis in various speeches,
both coaches and researchers need to manually select
and digest speech examples of their interests. 
\xingbo{It is ineffective and challenging for them to analyze humor in terms of both speech content and vocal delivery.}
They valued our attempt to build an interactive tool that systematically organizes these computational \xbRev{multimodal} features and provides concrete examples \xingbo{to verify the existing humor theories or reveal new insights into humor.}
We distilled the corresponding design requirements in Sec.~\ref{subsec:designrequirements}.


\subsection{Design Requirements}
\label{subsec:designrequirements}
According to the whole-to-part analysis  \xingbo{regarding the language strata},
our analytical system should support the hierarchical exploration of humor at different levels---speech level, context level, and sentence level.
We summarize the design requirements on these levels as follows.

\textbf{R1: Analyze text and audio simultaneously to reveal their correlations.} Our experts confirmed that both speech content and vocal delivery are considered necessary for a humorous effect. 
\xingbo{It is difficult to capture both of them by watching the videos.}
Therefore, at each level, the system needs to present textual and audio features concurrently to help users reason about the effective use of words and voice.

\textbf{R2: Visualize a speech level overview that shows vocal and verbal styles of humor, as well as their distribution.}
At the \textbf{speech level}, the system should summarize 
\xingbo{the timing of humor-related properties}---the humor is injected how frequently, under what condition (Or, what topic flow), to which part of the \xingbo{speech}, and with what verbal and vocal styles.
The visual summary of these properties
serves as guidance and should help users find specific humor snippets within a \xingbo{speech}.
For example, a communication coach might prioritize the very first humorous punchline (\emph{when}), 
to show students how to provide an impressive opening (\emph{objective}) by making small talks or sharing personal lives (\eg, \example{My brother's in prison.}).
Besides, as suggested by the experts, the visual summary of the humor distribution
should be integrated with temporal information, along with the topic flow and verbal feature statistics.

\textbf{R3: Provide a context-level overview that shows build-up elements of humor, as well as their relationships.}
Once zoomed in to a specific snippet, \textbf{context level} exploration is necessary for evaluating how a humorous story is written (\eg, how the key concepts in the punchline are first introduced and how they connect the pieces of humor stories), as well as a summary of delivery skills that are frequently used to help convey the story.
Both researchers and communication coaches
\xbRev{viewed} the 
contextual analysis of humor build-ups to be the most demanding.
Therefore, our system should primarily support users at this level.

\textbf{R4: Highlight the pairing of individual content words and humor-related verbal delivery units.}
We need to expose the co-occurrence between textual and audio features within each \textbf{individual sentence}, so to 
demonstrate the humor strategies with relevant concrete examples (\eg, words and utterance).
Within a snippet, the punchline is the most important sentence since it immediately triggers laughter.

\textbf{R5: Support intuitive interactions for helping users traverse among different levels, and reveal the different level of details}.
For example, 
our communication coaches suggested that the original audio and scripts of the speech should also be included in the system, in addition to a visual summary of humor.
The system should allow users to rapidly locate the segments of interest in the speech and playback the corresponding audio clips.

\begin{figure*}
\centering
\includegraphics[width=1.95\columnwidth]{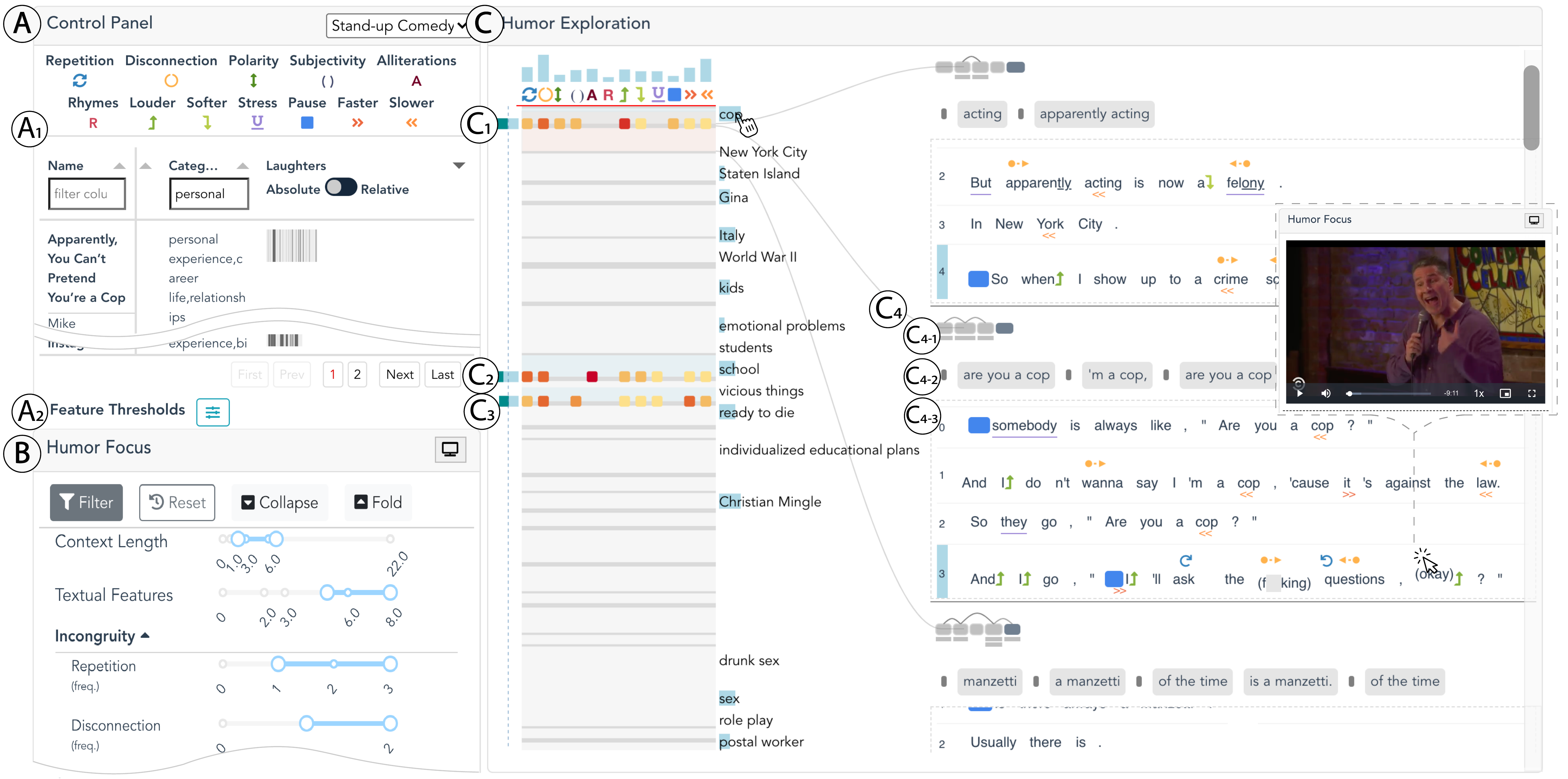}
\vspace{-3mm}
\caption{Interactive exploration of speech content and its vocal delivery of humor using DeHumor.
  The user uses {\cp} (A) to find a speech of interest.
  The {\hf} (B) helps the user to further narrow down to the humor snippets with certain verbal and vocal styles.
  The {\hx} (C) guides the multi-level exploration through an augmented time matrix (on the left) for summarizing humor features and context linking graphs (on the right) for analyzing humor context and its speech content and vocal delivery patterns.
  \xingbo{The user can click on the sentences to show and play the orginal video clips in the {\hf} (B).}}
\label{fig:teaser}
\vspace{-4mm}
\end{figure*}
\section{System Overview}
Motivated by the design requirements, we design 
and implement 
an interactive visual analytics system, {\name} (Fig.~\ref{fig:teaser}), to explore and analyze verbal humor in public speaking.
It combines 
multimodal humor features with visualization to facilitate users with insights into writing and delivering humor at three levels: speech level, context level, and sentence level.


{\name} contains three major modules: a data processing module, an analytics module, and a visual interface.
The \textbf{processing module} \xingbo{extracts humor snippets with aligned audio and transcripts from raw speech videos}
\xbRev{to}
support \xbRev{multimodal} analysis at different granularities.
The \textbf{analytics module} computes \xbRev{multimodal} humor features from audio and text, which characterize abstract and complex humor behaviors quantitatively.
The \textbf{interface} visualizes the features extracted by the \textbf{analytics module} to support intuitive exploration.
Here, we describe 
our processing module and then provide a brief overview of the interface. 
We delay the feature extraction in the analytical module, as well as their visual encodings, to \autoref{subsec:humorFeatureExtraction}.

\subsection{Data Preprocessing}

\label{subsec:process}
\textbf{Data collection.}
Given a humorous speech, we collect four kinds of data from it:
(1) We collect the meta-information (\eg,~title, speakers, and categories) for indexing and querying a specific \xingbo{speech}, so to enhance the usability of \name; 
(2) We label humor occurrence within a \xingbo{speech} based on the audience behavior markers 
(\ie, \texttt{[LAUGHTER]}) that are annotated in the transcripts; 
Previous studies have verified that laughter can reliably indicate humor ~\cite{morreall1986philosophy,nash2014language,bertero2016multimodal,hasan2019ur}; 
(3) We use the \emph{transcripts} for content analysis, and 
(4) the \emph{audio sequences} for verbal delivery analysis.

For demonstration purpose, we collect two speech datasets from \emph{TED Talks}~\footnote{\url{https://www.ted.com/talks}} and \emph{Comedy Central Stand-up}~\footnote{\url{http://www.cc.com/shows/stand-up}}, which will be described in detail in \autoref{sec:evaluation}. Users can prepare speech datasets of similar structures for their interests.

\textbf{Preprocessing.}
We process the collected data such that (1) the text script and audio are aligned to support multimodal analysis, and (2) the full speech data is segmented into \emph{humor snippets} to support context and sentence-level analysis.
To achieve the alignment, we first detect each word's starting time and ending time in the transcript using P2FA~\cite{yuan2008speaker}.
Thereafter, we align the audio and text modality together at the word level.
As for humor snippet segmentation, we regard a sentence immediately before a laughter marker as a \textbf{punchline} (\ie, the most important sentence that triggers the audience response).
We treat all of the sentences between two punchlines as the \textbf{candidate context paragraph} for the second punchline.
The intuition is that all the information that occurs after a punchline are potentially useful for \xbRev{building up} the next punchline. More concrete context recognition (shown in Sec.~\ref{subsubsec:context_analysis}) should come from these candidate sentences.
As a result, we split the transcripts at laughter markers. Each resulting \textbf{humor snippet} contains exactly one punchline (\ie, the last sentence of the segment) and its contexts (all the preceding sentences).
The audio is clipped correspondingly through the starting and ending times of the sentences.
Eventually, we organize the raw speech data into aligned audio and transcripts per snippet, per sentence, and per word.

\subsection{Interface Overview}
The user interface follows an overview-to-detail flow. 
In a collapsible {\cp} (Fig.~\ref{fig:teaser}{A}), a user can use the metadata (name, views, etc.) and the temporal distribution of punchlines (visualized as a \textbf{bar code chart} (in Fig.~\ref{fig:teaser}A{1})) to find their \xingbo{speech-of-interest}, which will be loaded in the main component, {\hx} view (Fig.~\ref{fig:teaser}C).
{\Hx} visualizes the humor-related details of a speech at different granularity. 
First, an augmented time matrix (on the left) summarizes the overall patterns of speech topics, humor insertion, and vocal delivery (\textit{\textbf{R1, R2}}).
With queries on the time matrix (\textit{\textbf{R5}}) or in the {\hf} (Fig.~\ref{fig:teaser}B), a user can locate a specific humor-snippet-of-interest into the context panel (on the right of Fig.~\ref{fig:teaser}{C}).
Within each snippet,
the user can examine the humor context (\textit{\textbf{R3}}) through the context linking graph, and understand the interactions between text and audio through the inline humor feature annotations among the transcripts (\textit{\textbf{R4}}).
Additional interactions on finding specific context links, related queries, etc. would further support users' exploration experience (\textit{\textbf{R5}}).

\section{DeHumor}
We describe the {\hx} view of {\name} in a bottom-up manner (Fig.~\ref{fig:teaser}C).
First, we illustrate the extraction and encoding of computational humor features in the \textbf{sentences and contexts}.
Then, the visual summary of the \textbf{whole speech}, as well as interactive features of the system, will be explained in detail.

\subsection{Humor Feature Analysis and Encoding}
\label{subsec:humorFeatureExtraction}
We utilize computational humor features to guide and enhance users' reasoning about the styles of verbal humor.
First, we describe how the textual (Sec.~\ref{subsub:text}) and audio (Sec.~\ref{subsub:audio}) features in Tab.~\ref{table:featuretable} are defined, computed, and represented at the \textbf{sentence level} (\textit{\textbf{R2}}).
To emphasize their potential co-occurrence, we encode all the features with inline glyphs (\textit{\textbf{R1, R4}}).
To further facilitate the \textbf{context} analysis (\textit{\textbf{R3}}), we design tools for extracting and visualizing the relationship among humor build-ups (Sec.~\ref{subsubsec:context_analysis}).

\subsubsection{Language Features and Glyphs}
\label{subsub:text}
We compute and encode three types of semantic features at the sentence level (\textit{\textbf{R4}}): incongruity, sentiment, and phonetics.
For each feature, a meaningful threshold is used to identify important words or phrases in the sentence, which are annotated with intuitive glyphs. These thresholds can be interactively adjusted by users according to the feature distribution in Fig.~\ref{fig:teaser}A2.

\begin{figure}[!htb]
  \centering
  \vspace{-3mm}
  \includegraphics[width=1\columnwidth]{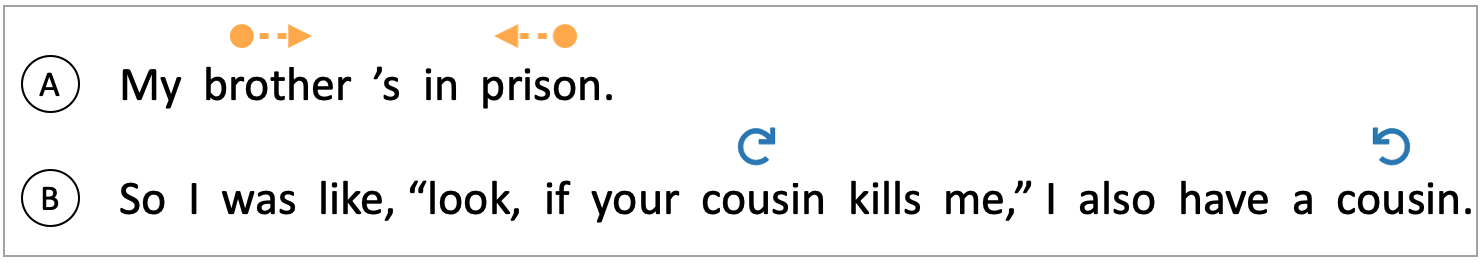}
  \vspace{-5mm}
  \caption{Examples of incongruity: (A) Disconnection (at beginning of a speech) and
  (B) \xingbo{Intra-sentence} repetition.
  }
  \vspace{-3mm}
  \label{fig:incongruity}
\end{figure}

\textbf{Incongruity.}
Contrasting incongruous concepts (e.g., ``clean desk'' and ``cluttered desk drawer''~\cite{mihalcea2005making}) is classic for achieving the comic effect.
The semantic incongruity of a sentence 
can be modeled by the repetition \xbRev{and} disconnection, or the relative semantic similarities between word pairs~\cite{yang2015humor}.
\textbf{Disconnection} captures the least semantically similar word pair in the sentence. 
As shown in Fig.~\ref{fig:incongruity}A, a pair of dashed arrows (\includegraphics[height=\fontcharht\font`\B]{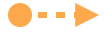}, \includegraphics[height=\fontcharht\font`\B]{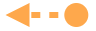})
are placed above the words ``brother'' and ``prison'' to indicate their disconnection.
In contrast, \textbf{\xingbo{intra-sentence} repetition} focuses on the most similar pair.
In Fig.~\ref{fig:incongruity}B, a pair of curved arrows
(\textcolor{repetition}{\reflectbox{\faUndo}}, \textcolor{repetition}{\rotatebox[origin=c]{360}{\faUndo}}) show the repetition of the two ``cousin''s.
The arrows in a pair are pointed to each other, showing the sequential orders and positions of the corresponding word pair in the sentence.
We calculate semantic similarity using the cosine distance on the GloVe~\cite{pennington2014glove} embedding. 
At the sentence-level, we also annotate the sentences that have word pairs with strong disconnections (\textcolor{disconnection}{\faCircleONotch}) or repetitions (\textcolor{repetition}{\faRefresh}).

\begin{figure}[!htb]
  \vspace{-3mm}
  \centering
  \includegraphics[width=\columnwidth]{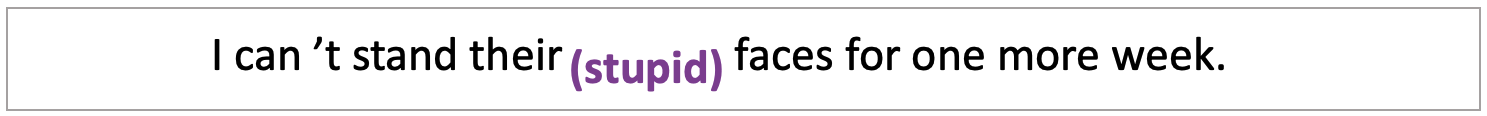}
  \vspace{-7mm}
  \caption{An example of sentiment expression.}
  \vspace{-2mm}
  \label{fig:sentiment}
\end{figure}

\textbf{Sentiment.}
Expressing strong sentiment using polarized expressions (how emotionally positive and negative) and subjective statements (how personal) enables a speaker to empathize with the audience. 
The \textbf{polarity} includes both the sentiment direction and \xbRev{sentiment intensity}.
We use vertical offsets to indicate words with strong polarity. 
For example, ``stupid'' in Fig.~\ref{fig:sentiment} has a negative polarity, and is therefore displayed with a downside vertical offset.
The \textbf{subjectivity}, on the other hand, is shown with brackets ``\textcolor{subjectivity}{\textbf{( )}}'' around a word.
As shown in Fig.~\ref{fig:sentiment}, ``stupid'' is associated with the speaker's subjective opinion.
We measure word-level and sentence-level polarity and subjectivity using the resource of word annotations and clues for sentiment in \cite{wilson2005recognizing}.

\textbf{Phonetic style.}
Phonetic style is often used to achieve catchy verbal deliveries, making the comic effect more memorable and engaging~\cite{mihalcea2005making}.
The most common techniques include (1) \textbf{alliteration chains}, which denote multiple words that begin with the same phones, and (2) \textbf{rhyme chains}, which include words ending with the same syllables.
We utilize the CMU Pronouncing Dictionary~\footnote{\url{http://www.speech.cs.cmu.edu/cgi-bin/cmudict}} within every sentence, and visually underline the corresponding characters that are responsible for creating the chain.
\begin{figure}[!htb]
  \centering
  \vspace{-3mm}
  \includegraphics[width=\columnwidth]{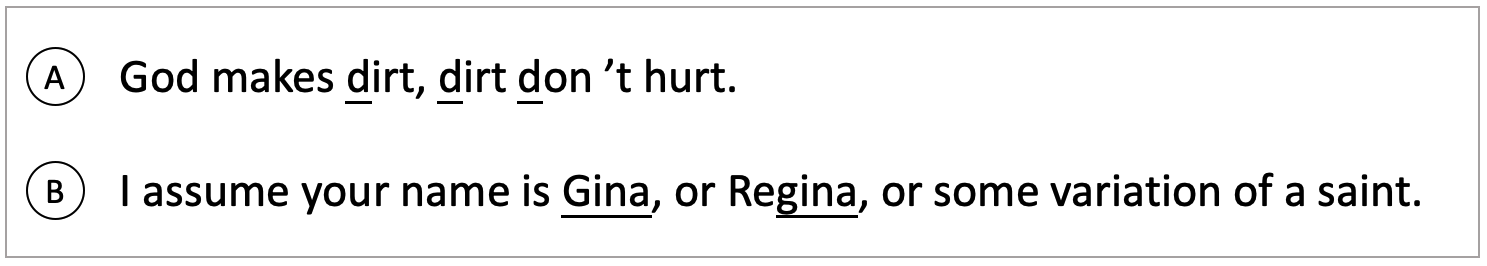}
  \vspace{-7mm}
  \caption{Examples of phonetic styles.
  A: Alliteration.
  B: Rhyme.}
  \vspace{-5mm}
  \label{fig:phonetic}
\end{figure}

\subsubsection{Audio Features and Glyphs}
\label{subsub:audio}

We extract and encode the following four representative audio features that reveal the speaker's vocal delivery style.
Most of these features are captured by computing its relative significance within a sentence or paragraph.
For example, in the speed variation below, we define words to be significantly faster, if it is $N$ times faster than the average of the speed in a sentence, or $M$ times of standard deviations away from the average, with given thresholds $N$ and $M$ (defaulted to $1$ and $1.5$ respectively).

\begin{figure}[!htb]
  \centering
   \vspace{-3mm}
  \includegraphics[width=\columnwidth]{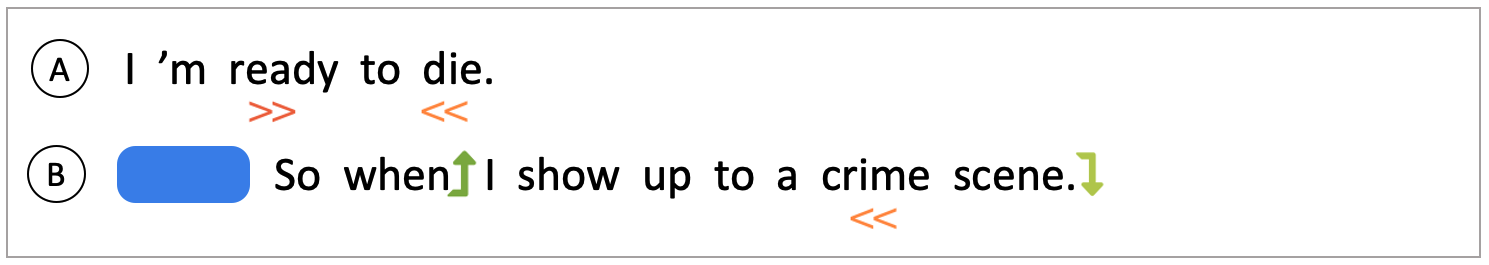}
  \vspace{-7mm}
  \caption{Examples of vocal delivery styles. A: Speed variations. B: A combination of pause, and volume and speed variations.}
  \label{fig:audiofeature}
  \vspace{-2mm}
\end{figure}

\textbf{Speed variation.} We compute the Syllables Per Minute (SPM) for each word, 
as well as the average and standard deviation (SD) of SPM for each sentence.
As shown in Fig.~\ref{fig:audiofeature}A, we encode the words which are significantly faster than the sentence as ``faster~(\textcolor{fast}{\faAngleDoubleRight})''.
Similarly, the words which are significantly slower 
will be 
labeled as ``slower~(\textcolor{slow}{\faAngleDoubleLeft})''.

\textbf{Pause.} We calculate the time intervals between two words. If the interval exceeds a threshold (that defaults to 0.5s), a dark blue rectangle will be drawn in front the corresponding word (\eg, ``So'' in Fig.~\ref{fig:audiofeature}B). The width of the rectangle encodes the pause length.

\textbf{Volume variation.} 
We mark the words that are significantly louder or softer than the preceding word in the sentence.
They are labeled as ``louder (\textcolor{louder}{\faLevelUp})'' or 
``softer (\textcolor{softer}{\faLevelDown})''.

\textbf{Pitch stress.}
Similar to the volume variation, 
we 
derive
words that are significantly higher pitched or have more pitch variation based on the pitch contours, and encode them with ``stress (\textcolor{stress}{\underline{\textbf{U}}})''.

The thresholds above are set according to \cite{rubin2015capture, wang2020voicecoach}.
They are fine-tuned empirically by testing on audio samples of speech data
and can be interactively adjusted in Fig.~\ref{fig:teaser}A.

\subsubsection{Humor Context Analysis and Linking}
\label{subsubsec:context_analysis}

To reveal the relationship among build-ups of a punchline (\textit{\textbf{R3}}), 
we extract and link similar concepts in the punchline context.
As mentioned in Sec.~\ref{subsec:designrequirements}, a speaker would more frequently repeat useful concepts to help prepare the audience for the upcoming punchline.
In the example below, 
the core takeaway of the punchline is that Germany \emph{does not} have \example{fantastic food}. 
The message becomes clear because of several repetitions in preceding lines.
First, the speaker emphasizes his/her focus on the two countries by repeating (\example{the Italian community}, \example{Their people}, \example{Italy}) and (\example{Germany}) in several places.
Second, in Lines \#3 and \#5, the different modifiers \example{a} and \example{no} before the repeated \example{stigma for (being) evil} highlights the opposite reputations of Germany and Italy after WWII, and therefore builds a natural comparison between the two.
The comparison is then carried on to the punchline, implying the German food is the opposite of Italian's.

\qbox{
Let me go after \gitaly{the Italian community}.\\
\gitaly{Their people} get off easy.\\
\ggermany{Germany} \gstigma{has a stigma} \gevil{for being evil}.\\
But if you check history, \gitaly{Italy} \ggermany{fought right alongside Germany} in WWII.\\
But we \gstigma{have no stigma} \gevil{for evil}, and do you guys know why?\\
It's because we have fantastic food.
}

With this example, we first present an algorithm that captures such \xingbo{inter-sentence} repetitions and then describe the visual display.


\paragraph*{\textbf{Concept Grouping Algorithm}}
Concept grouping may sound trivial at first glance---Naive string match among different tokens may suffice if we assume concepts are always repeated in strictly identical forms.
However, in practice, we frequently observe context rephrasing.
For example, while the concept entity \example{Italy} is introduced as a modifier for its community in the first sentence, in Line \#2 it is just implicitly referred by a pronoun.
Beyond entities, more concepts appear in the form of modifier segments (synonym adjectives, similar prepositions on different entities, etc.), just like \example{for being evil} and \example{for evil} in Lines \xbRev{\#3 and \#5} respectively.
Another intuitive method---grouping semantically similar full sentences---can relax the constraint of ``identical repetition'', but is likely to miss cases when only a small part of the sentences have overlapping concepts (\eg, \example{Germany} in Lines \#3 and \#4).

\begin{figure}[!htb]
  \vspace{-0.1in}
  \centering
  \includegraphics[width=\columnwidth]{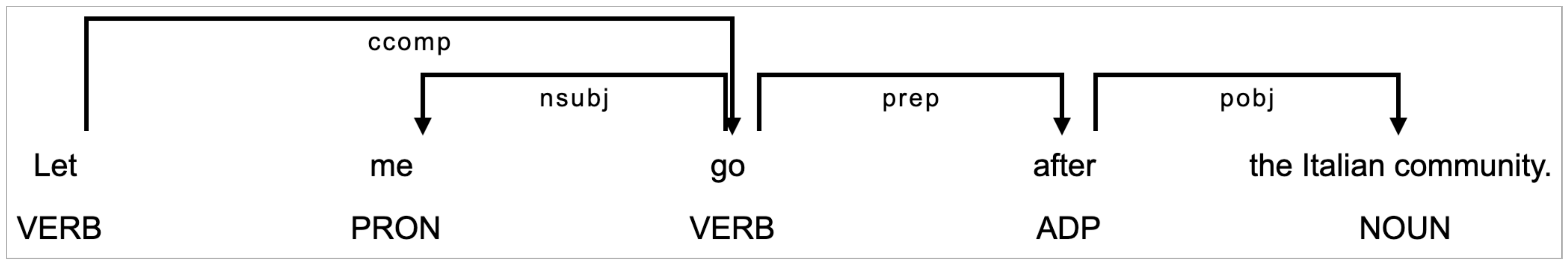}
  \vspace{-0.2in}
  \caption{The dependency tree of Sentence \#1 in the Sec.~\ref{subsubsec:context_analysis} example.
  }
  \label{fig:parse_tree}
  \vspace{-0.1in}
\end{figure}

To capture free-form concepts hidden within full sentences, our grouping method performs two crucial steps:
First, \textbf{to separate concepts from long sentences}, we induce subphrases by traversing the dependency tree of a given full sentence\footnote{With the NLP processing library SpaCy: \url{https://spacy.io/}}. 
For example, with Line 1 parsed into Fig.~\ref{fig:parse_tree}, we get verb phrases like \example{go after the Italian community} as well as noun phrases like \example{the Italian community}.
Second, \textbf{to merge the rephrasings}, we perform density-based clustering on the induced subphrases based on their semantic similarity:
we resolve coreferences between sentences (\eg, \example{Their people} in Line \#2 becomes \example{Italian people})\footnote{With \url{https://github.com/huggingface/neuralcoref}}.
Then, we transform phrases into feature vectors with a state-of-the-art universal sentence encoder~\cite{reimers-2019-sentence-bert} and then compute the cosine similarity in the embedding space. \xingbo{This approach is effective for semantic textual similarity (STS) task (with an accuracy score of 85\% on STS Benchmark).}
Finally, because suphrases recognized through the parsing tree overlap with each other, we reduce redundancy in the extracted repetitions by keeping the segment with the largest possible similarity with its cluster (\eg, in \example{go after the Italian community}, the first two words are considered unnecessary.)

\begin{figure}[t]
  \centering
  \vspace{-3mm}
  \includegraphics[width=\columnwidth]{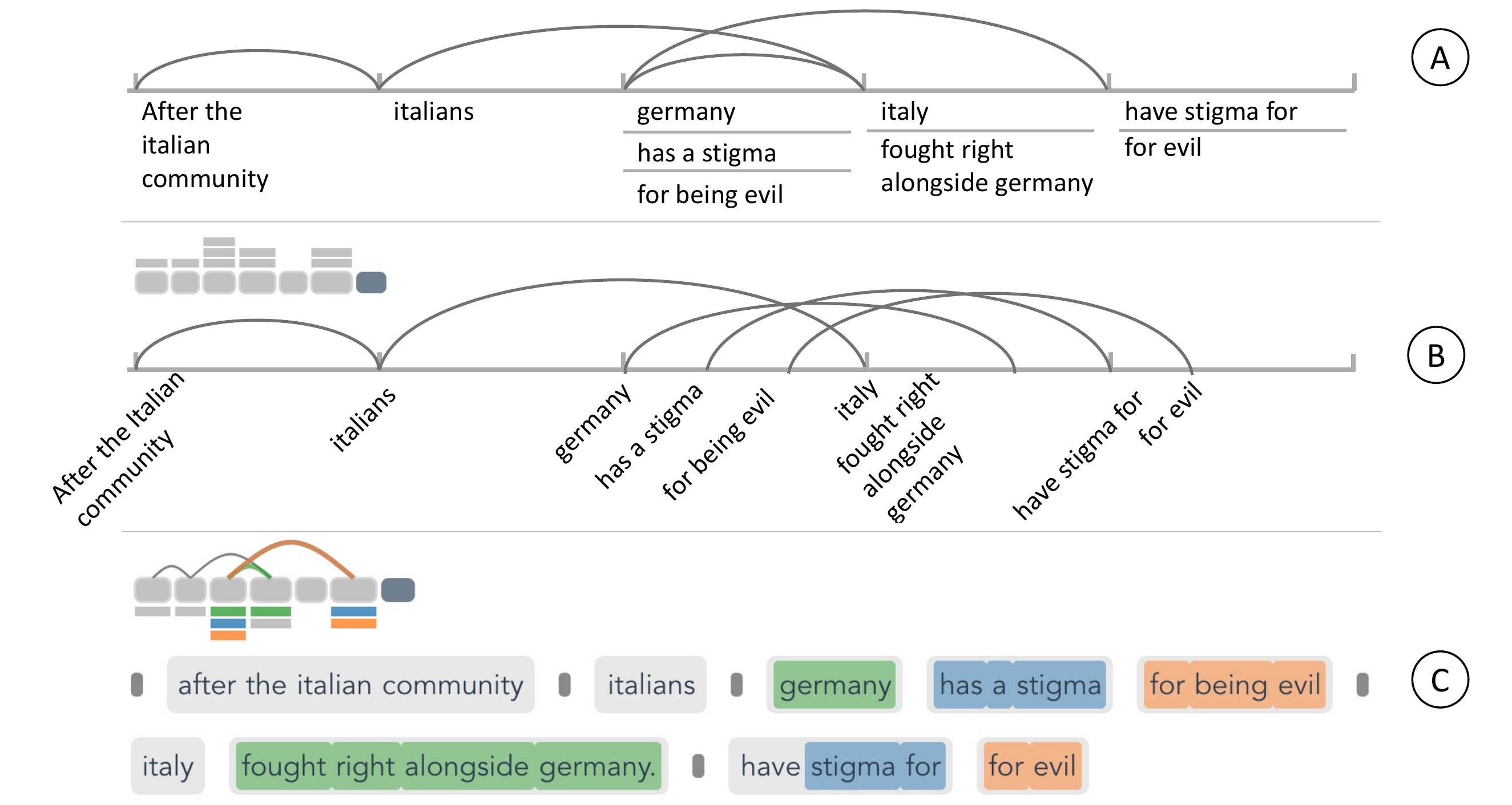}
  \vspace{-6mm}
  \caption{Alternative designs for context linking.
  Compared to our current design (C), A and B are less space-effective and more cluttered.}
  \label{fig:contextalternatives}
  \vspace{-5mm}
\end{figure}

\paragraph*{\textbf{Visualizing Contextual Repetitions}}
We design a context linking graph to display the extracted \xingbo{inter-sentence repetition} occurrences.
As shown in Fig.~\ref{fig:contextalternatives}C,
(or a more complete version can be seen in Fig.~\ref{fig:comedy-cousin_case}), 
the graph follows a three-stage design, such that it gradually reveals the concrete context information to the user and traverse from the context-level (\textbf{\textit{R2}}) to the sentence-level (\textbf{\textit{R3}}).

The graph first provides a \textbf{context distribution summary}, which shows \emph{how} 
\xbRev{the sentences are}
connected to each other through repeated concepts.
A rounded gray rectangle represents a sentence, \xbRev{and its} \xingbo{horizontal length} encodes the sentence length.
We highlight the most important punchline with a denser gray color (\textbf{\textit{R4}}).
We connect rectangles with arc links if their corresponding sentences share repetitive concepts, and add thin lines below the rectangles to denote the presence of these concept.
The combination of the links and the lines help highlight different structures of build-up for humor.
For example, Fig.~\ref{fig:contextalternatives}C implies that most repetitions occur in the first half of the context, especially the third sentence, with three concepts repeated elsewhere.
There is no link between the punchline and the context, suggesting that the punchline is disconnected from the previously mentioned ideas.
Then, it presents the concrete \textbf{repetitive concepts} to show \emph{what} are used for building up the context, but still omits other details in the related sentences.
These concepts are sorted based on the order they occur in the text segment, and each small dark rectangle marks the boundary for a sentence.
Finally, the repetitive concept helps locate \textbf{detailed sentence} contents and their associated humor features line by line, such that the abstract concepts can be integrated with the complete story.

\xbRev{We also design a set of interactions to enable the traversal of the context summary, repetitive concepts, and detailed sentences.}
Specifically, when users hover over a rectangle (sentence) of interest in the context summary,
all its connections and the corresponding groups of repetitive concepts will be highlighted in different colors.
Conversely, as users hover over a phrase in the repetitive concepts, its repeated concepts in other sentences and the corresponding links in the context summary will be highlighted.
To facilitate more insights into word usage and verbal delivery, 
we support a quick reference to the original content in the sentence when users click in context summary or on a specific phrase.

\textbf{Design alternatives.}
We discuss the trade-offs of alternative designs for the context summary and repetitive concepts in our iterations.
Initially, we attempted to combine the links with concrete concepts.
In Fig.~\ref{fig:contextalternatives}A, each tick in the horizontal axis marks the corresponding sentence.
Below the tick, repetitive concepts are listed vertically according to the order of their occurrences.
While this design does not require separating concepts from the links, this layout could easily exhaust
\xbRev{the available horizontal or vertical space when we have a long context or a large number of repeated concepts.}
It also sacrifices the temporal ordering of the concept occurrence and makes the concept exploration less intuitive. 
We then tried to place repetitive concepts slantingly along the axis according to their occurrence ordering, and directly link the repeated concepts.
Because the notion of ``sentence separation'' becomes less apparent, we further add a repetition distribution on the top, such that users can count the repeated concepts.
The design (Fig.~\ref{fig:contextalternatives}B) saves the space and recovers temporal information, but 
\xbRev{causes}
visual cluttering issue. 
Specifically, when the number of concepts increases, linking concepts---instead of their corresponding sentences like in Fig.~\ref{fig:contextalternatives}A---induces 
additional overhead for distinguishing the intertwined links.
That said, the concept of overview-to-detail was favored in some preliminary discussions with end-users.
Thus, we thought short links among sentence glyphs and concepts overflowed among multiple lines would create the least cognitive load and would be the most space-efficient---which is exactly our design in Fig.~\ref{fig:contextalternatives}C.

\subsection{Augmented Time Matrix}
\label{subsec:augmented_timematrix}
Besides sentence- and context-level, we design an augmented time matrix that provides an overview of \xingbo{distribution of humor occurrences and the related features of speech content and vocal delivery
at the \textbf{speech level} (\textit{\textbf{R1, R2}}).}

\begin{figure}[t]
  \centering
  \includegraphics[width=0.9\columnwidth]{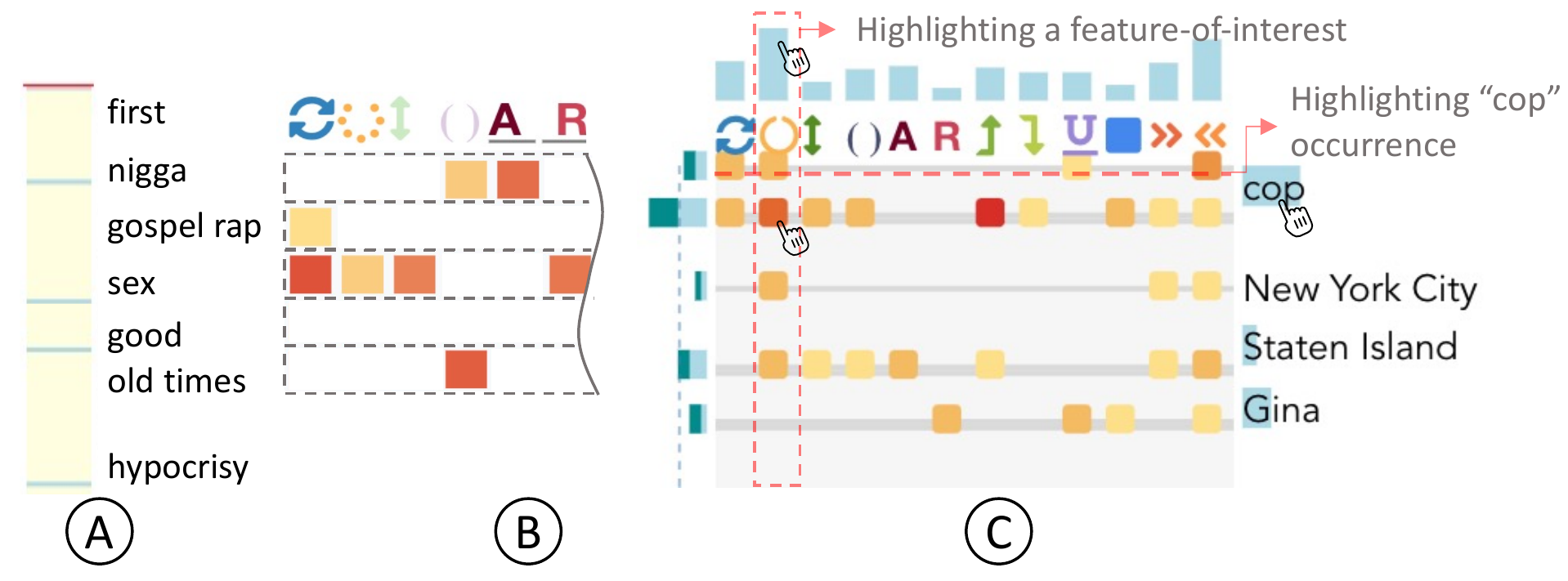}
  \vspace{-4mm}
  \caption{\xingbo{Alternative designs for speech summary using an annotated barcode chart (A) and a matrix summary of humor features (B). Our current augmented time matrix design (C) combines (A) and (B) to summarize the timing of humor and the distribution of humor features. Users can hover over a feature or a keyword to highlight its occurrences.}}
  \vspace{-4mm}
  \label{fig:summaryalternatives}
\end{figure}

\xingbo{As shown in Fig.~\ref{fig:summaryalternatives}C, the barcode chart of the time matrix shows the humor distribution.}
The big gray rectangle shows the whole time matrix from the top to the down.
The darker horizontal lines in the time matrix indicate timestamps where the punchlines occurred.
Therefore, by definition of the humor snippet (Sec.~\ref{subsec:process}), the light gray area between two horizontal lines \xbRev{indicates} the context length between punchlines.
\xingbo{If the time intervals between punchlines are too small, the horizontal gray lines (\ie, punchlines) are merged into one thick line to reduce visual clutter.}

\xingbo{Besides, we organize the humor features, including the word usage, vocal delivery, and key concepts, around the matrix to summarize 
their distribution for each punchline and across different punchlines.}
A bar chart is placed at the top to show the total occurrences of humor features in the punchlines, where each bar represents a feature, and the height of the bar indicates the feature frequency.
Then, for each punchline, a stacked bar is placed on the left at the same vertical position, where the dark green bar reveals the frequency of textual features, and the light green bar reveals the frequency of audio features.
Moreover, colored boxes on the punchlines (dark gray lines) imply the frequencies of humor features.
The darker the color, the higher the frequency.
To reveal the key concepts for humor snippets, 
we extract keywords for each snippet using TextRank~\cite{mihalcea2004textrank},
and place them along the time matrix in temporal order. 
A horizontal blue bar is overlaid to denote the frequency of the keywords.
\xingbo{Users can hover over a feature-of-interest (\ie, a bar at the top or a colored box in the matrix) or a keyword to highlight its occurrences across the whole speech in the time matrix.
When a user clicks a keyword or in the time matrix, 
the system will
highlight the corresponding
punchline and its context in the context linking graph.} 

\textbf{Design alternatives.}
Initially, we have considered separating the humor feature summary (Fig.~\ref{fig:summaryalternatives}B) from the content summary (Fig.~\ref{fig:summaryalternatives}A).
However, the sparse feature matrix takes up a large space and do not provide any context information about the punchline. Particularly, some of our end-users complained that it is hard to figure out where the corresponding punchlines when exploring the feature summary.
Thus, in our current design, we integrate the feature summary into the timeline to enhance both the temporal and contextual information for humor features.

\subsection{Interactions}
Our system supports a rich set of interactions to ease the multi-level exploration of humor \textbf{\textit{(R5)}}. 

\textbf{Details-on-demand through clicking.}
Once a user clicks a \xingbo{speech} of interest in the {\cp}, 
the {\hx} will be updated accordingly.
When the user clicks on a keyword or in the augmented time matrix,
the corresponding humor snippet will be scrolled to the top in the content exploration, and the corresponding sentence in the context linking graph and the transcript will be highlighted.

\textbf{Linked scrolling.}
When users scroll in the content exploration, the time range of the visible humor snippets will be highlighted in the augmented time matrix.

\textbf{Active query through searching, sorting, and filtering}.
Users can search a \xingbo{speech} or sort \xingbo{speeches} according to a specific criterion in the {\cp}.
Also, they can apply filters in the {\hf} to find different styles of punchlines. Then, the corresponding humor snippets will be highlighted in the {\hx} view.

\section{Evaluation}
\label{sec:evaluation}

\xbRevise{We demonstrate how \name{} helps users gain insights into \xbRev{the verbal content and vocal delivery of humor speeches} through two case studies and expert interviews. The experts include two humor researchers (\textit{E1}, \textit{E6}) and two communication coaches (\textit{C1}, \textit{C3}).
\textit{E1} and \textit{C1} have participated in the design process, while \textit{E6} and \textit{C3} were new to our system before the interviews.
Specifically, 
\textit{E6} holds a master degree in linguistics, and her research focuses on the pragmatics of humor. \textit{C3} has been a communication coach for four years. He is also a stand-up comedian and has performed at famous venues (\eg, Broadway).
During the interviews, the experts used Dehumor to explore two humor datasets, which consist of 157 shows of \textit{Comedy Central Stand-up} and 1,876 \textit{TED Talks}.
The cases in Sec.~\ref{subsec:case1} and Sec.~\ref{subsec:case2} were found by
\textit{E1} and \textit{C1}, respectively. All the experts’ feedback was 
collected and 
reported in Sec.~\ref{subsec:expert_interviews}.}

\xbRevise{To better illustrate the cases, 
we highlight the important humor analysis steps: \casetag{Context relationship analysis} for context exploration, \casetag{Humor context} and \casetag{Punchline} for humor description, and \casetag{Feature analysis} for punchline analysis.}

\subsection{Case Study on Stand-up Comedy}
\label{subsec:case1}
In this case, \textit{E1} used {\name} to explore the ``stand-up comedy'' dataset and check how comedians effectively set up humor about the funny incidents happened in their lives. In particular, she was interested in the word usage of punchlines and would like to see how it helps create humorous effects.
First, \textit{E1} filtered the speeches by keywords ``personal experience'' in the {\cp} (Fig.~\ref{fig:teaser}A1). Then, she felt interested in \xingbo{speeches} that frequently involve humorous moments. Thus, she sorted the speeches by the total occurrences of laughter and selected the first \xingbo{speech} named  ``Apparently  You  Can’t  Pretend  You’re  a  Cop''\footnote{Comedy video url: \url{https://bit.ly/3u1lcZq}}.

\textbf{\textit{Case Context}}: 
This case includes three 
speaker's personal experiences in the selected \xingbo{speech}.
(1) The speaker talked about his experience at a crime scene. 
The people there wanted to check if he was a cop. 
(2) The speaker told a story when he was a teacher. He was once threatened by a student during a fight. 
(3) Following the previous story, the speaker described that after the fight, both he and the student claimed that they were ready to die.

\vspace{-1mm}
\subsubsection{Overall styles of verbal humor}
\textit{E1} wondered what the major characteristics of this speaker's humor strategies in his speech are. By observing the height of bars at the top of the augmented time matrix (Fig.~\ref{fig:teaser}C), she found that ``repetition (\repetition)'' and ``disconnection (\disconnection)'' frequently occur in the punchlines. She was curious about \textit{what words the speaker used to create incongruity and how he delivered} (\textit{\textbf{R1, R4}}). 
As she skimmed through the dark gray lines in the time matrix, she found clusters of punchlines that are close to each other across the whole speech. She wondered \textit{how the speaker set up humor within a short context} (\textit{\textbf{R2, R3}}). To answer the two questions, she adjusted the filters of context length and textual features in the \textit{humor focus} (Fig.~\ref{fig:teaser}B). The snippets that satisfied the conditions were highlighted with colored feature statistics in the augmented time matrix (Fig.~\ref{fig:teaser}C). Next, she explored them in detail.

\vspace{-1mm}
\subsubsection{Digging into humorous snippets}
She found that the first highlighted snippet appeared at the beginning of the speech (Fig.~\ref{fig:teaser}C1), where the keyword ``cop'' was spotted (\textit{\textbf{R2}}). As she hovered over the word, its other occurrences were also marked in dashed red lines in the time matrix, one of which fell into the current snippet of her interest. 
Then, she clicked the dashed line to locate the corresponding snippet and its context linking graph to see the story development (\textit{\textbf{R1, R3, R5}}).
\casetag{Context relationship analysis}
Through the links (Fig.~\ref{fig:teaser}C4-1) and repeated phrases
(\example{are you a cop}, \example{'m a cop}, \example{are you a cop})
in the context distribution summary of the graph (Fig.~\ref{fig:teaser}C4-2) \textbf{(\textit{R4})}, she guessed that the speaker was having a conversation with someone else about the ``cop'' identity. She followed the link connections among sentences and observed the corresponding content (Sentences \#0 to \#2 in Fig.~\ref{fig:teaser}C4-3) (\textit{\textbf{R3}})---\casetag{Humor context} the speaker was asked by the people at a crime scene about whether he \xbRev{was} cop. Since he is not an actual cop, he did not want to explicitly make an illegal claim that he \xbRev{was} a cop.
\textit{E1} navigated to the punchline (Sentence \#3 in Fig.~\ref{fig:teaser}C4-3) to see how the speaker responded to the question.
\casetag{Punchline} She discovered that the speaker quoted a common trope for police in movies and TV (i.e., \example{I'm asking the f**king question!} in Fig.~\ref{fig:teaser}C5-3) and misled the people at the crime scene to believe that he \xbRev{was} a cop.
\casetag{Feature analysis} Specifically,
\textit{E1} referred to the feature annotations in the sentence, finding that the speaker raised his voice (\louder) on the first few words in the punchline (\example{And}, \example{I}).
Then the speaker paused a little bit (\pause) before revealing the essence of the content---``I will ask the f**king (\polarity, \subjectivity) questions''. Finally, he strengthened his annoyance by the previous question about his identity (\louder) through a tone particle ``okay''.
\textit{E1} concluded that the ``cop'' repetitions in the context and his voice modulation in the punchline renders the speaker's annoyance and enhance humorous effect.

\begin{figure}[t]
  \centering
  \includegraphics[width=0.95\columnwidth]{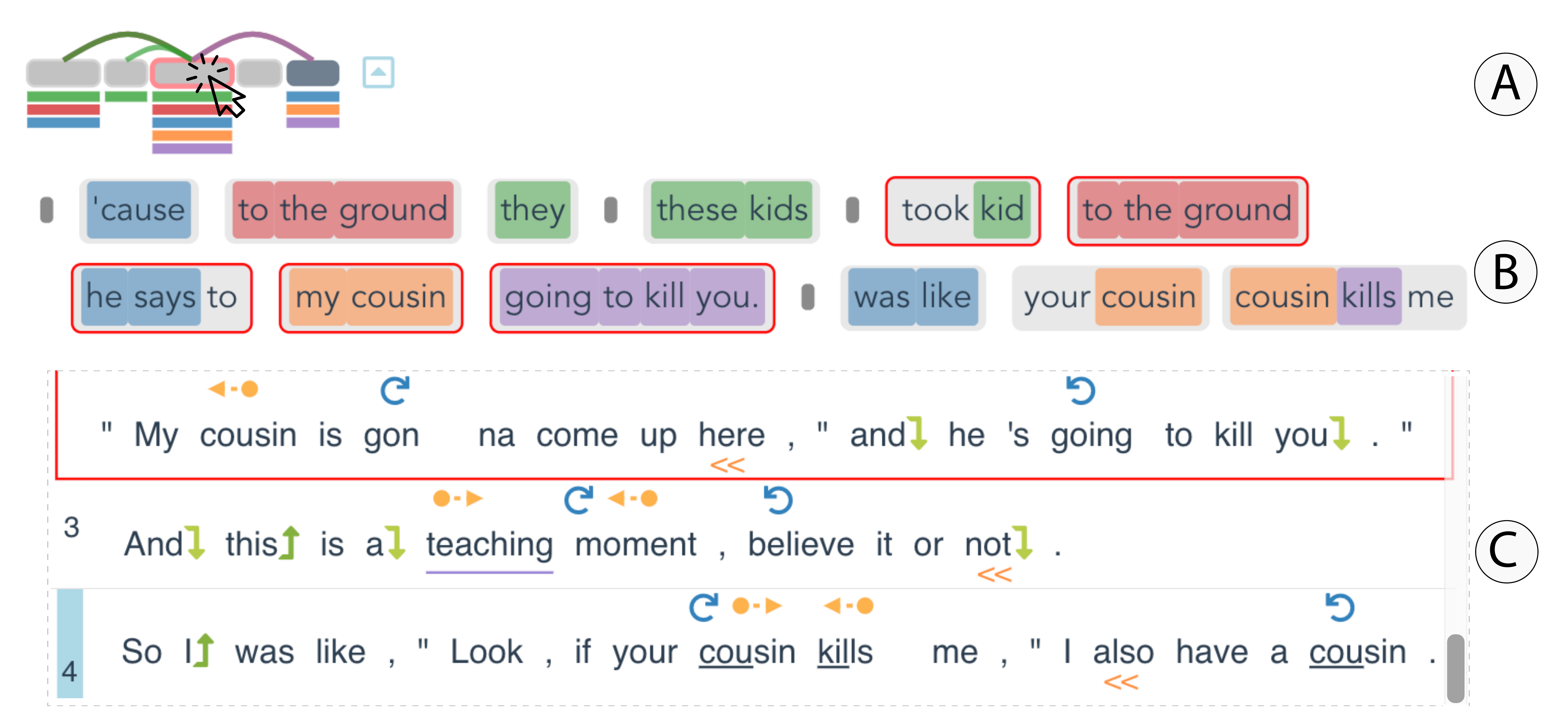}
  \vspace{-5mm}
  \caption{The context linking graph of the first snippet in Fig.~\ref{fig:teaser}C2, 
  with Sentence \#2, as well as corresponding repeated phrases and their links being highlighted.}
  \vspace{-5mm}
  \label{fig:comedy-cousin_case}
\end{figure}

Then, \textit{E1} clicked in the second highlighted snippet (\textit{\textbf{R5}}) (Fig.~\ref{fig:teaser}C2).
\casetag{Context relationship analysis}
She noticed that the third rectangle in the context summary (Fig.~\ref{fig:comedy-cousin_case}A) has the most bars attached below, suggesting it contains the most repetitive phrases. Then, she clicked on the rectangle and the corresponding repetitions were highlighted. By observing the red rectangles (\example{to the ground}) and purple rectangles (\example{going to kill you}, \example{kills}) of the repeated phrases (Fig.~\ref{fig:comedy-cousin_case}B),
she assumed there was a big fight. By following the links in the graph from beginning to end and exploring the content (Fig.~\ref{fig:comedy-cousin_case}A), \casetag{Humor context} she grasped that the speaker wrestled a student to the ground during a fight. Then, the student threatened that his cousin would come and help him kill the speaker. 
\casetag{Punchline \& Features}
The speaker said he also had a cousin (\repetition) (Sentence \# 4 in Fig.~\ref{fig:comedy-cousin_case}C) in response to the student, implying his cousin would also help him and  kill the student if the student's cousin killed him (\textit{\textbf{R4}}).

\begin{figure}[t]
  \centering
  \includegraphics[width=\columnwidth]{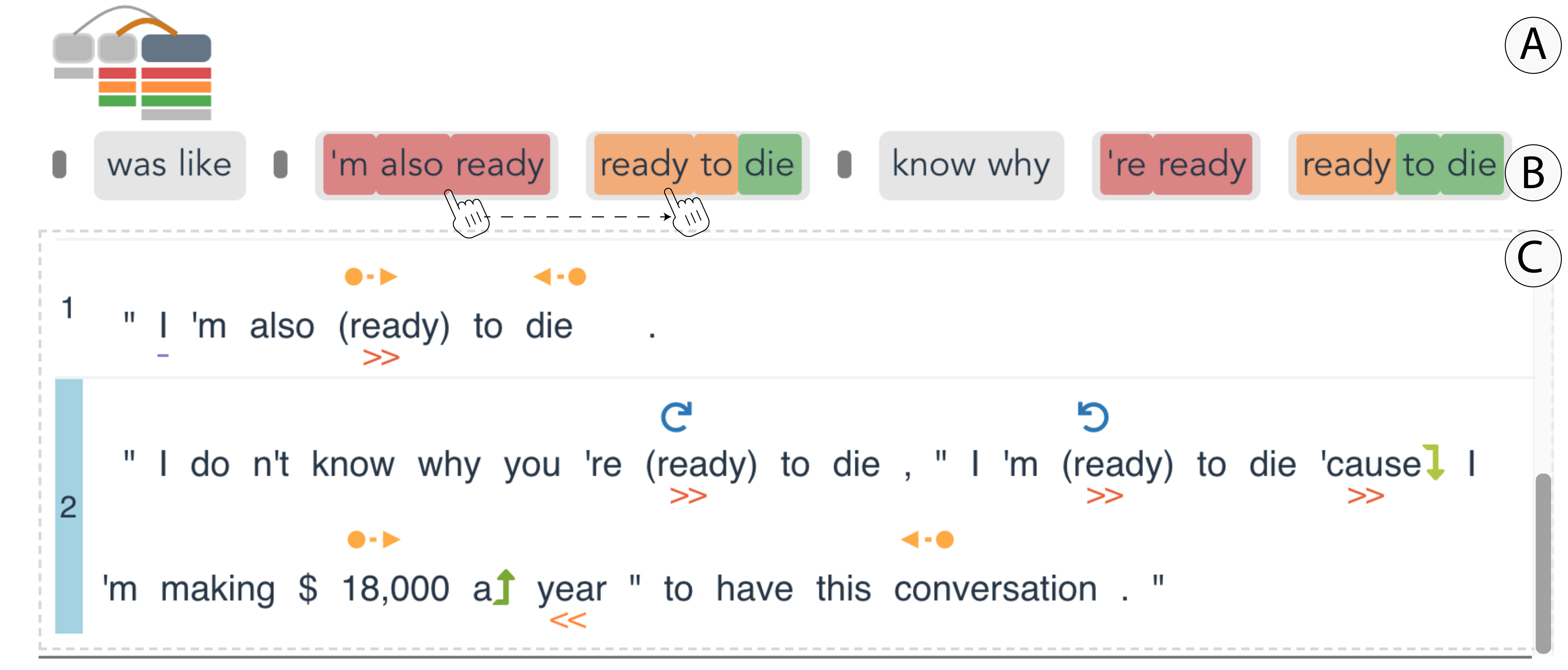}
  \vspace{-5mm}
  \caption{The context linking graph of the snippet in Fig.~\ref{fig:teaser}C3, with the repeated phrases from Sentence \#2 and their links being highlighted.}
  \label{fig:comedy-read-to-die_case}
  \vspace{-5mm}
\end{figure}

Then, \textit{E1} scrolled down until the third highlighted snippet (Fig.~\ref{fig:teaser}C3). 
\casetag{Humor context} She found that the speaker won the fight with the mentioned student. The student said that he was ``ready to die'' after losing the fight. 
Then in the current snippet (Fig.~\ref{fig:comedy-read-to-die_case}), \textit{E1} tracked the colored repeated phrases (``ready to die'') (Fig.~\ref{fig:comedy-read-to-die_case}B) (\textit{\textbf{R3}}). She
found that the speaker responded with ``I'm also ready to die'' and explained in the punchline (Sentence \#2 Fig.~\ref{fig:comedy-read-to-die_case}C) (\textit{\textbf{R4}})---\casetag{Punchline}
the speaker felt disappointed about arguing with the naughty student because he was paid extremely low wages at school to deal with such a big trouble maker (\ie, the student).
\casetag{Feature analysis}
Specifically, the labeled word pair ``\emph{18000}'' (\includegraphics[height=\fontcharht\font`\B]{pictures/discon0.png}) and ``\emph{conversation}'' (\includegraphics[height=\fontcharht\font`\B]{pictures/discon1.png}) in the punchline (Sentence \#2 in Fig.~\ref{fig:comedy-read-to-die_case}C) 
contrasts the low-paid job with the high effort of teaching the student.
In addition, the speaker even inserted pauses (\pause) and raised his voice (\louder) after \example{18000} to emphasize his complaints about his challenging but low-paid work.

As for takeaways of this exploration process, \textit{E1} concluded that the speaker set up conversation scenarios to narrate his interesting personal experiences. He is good at using \xingbo{contextual} repetitions to connect pieces of a story and using words to create incongruity. Moreover, he modulated his voice (\eg, using pauses and increasing volume) to express his emotion and strengthen the humor.

\begin{figure*}
\centering
\includegraphics[width=1.97\columnwidth]{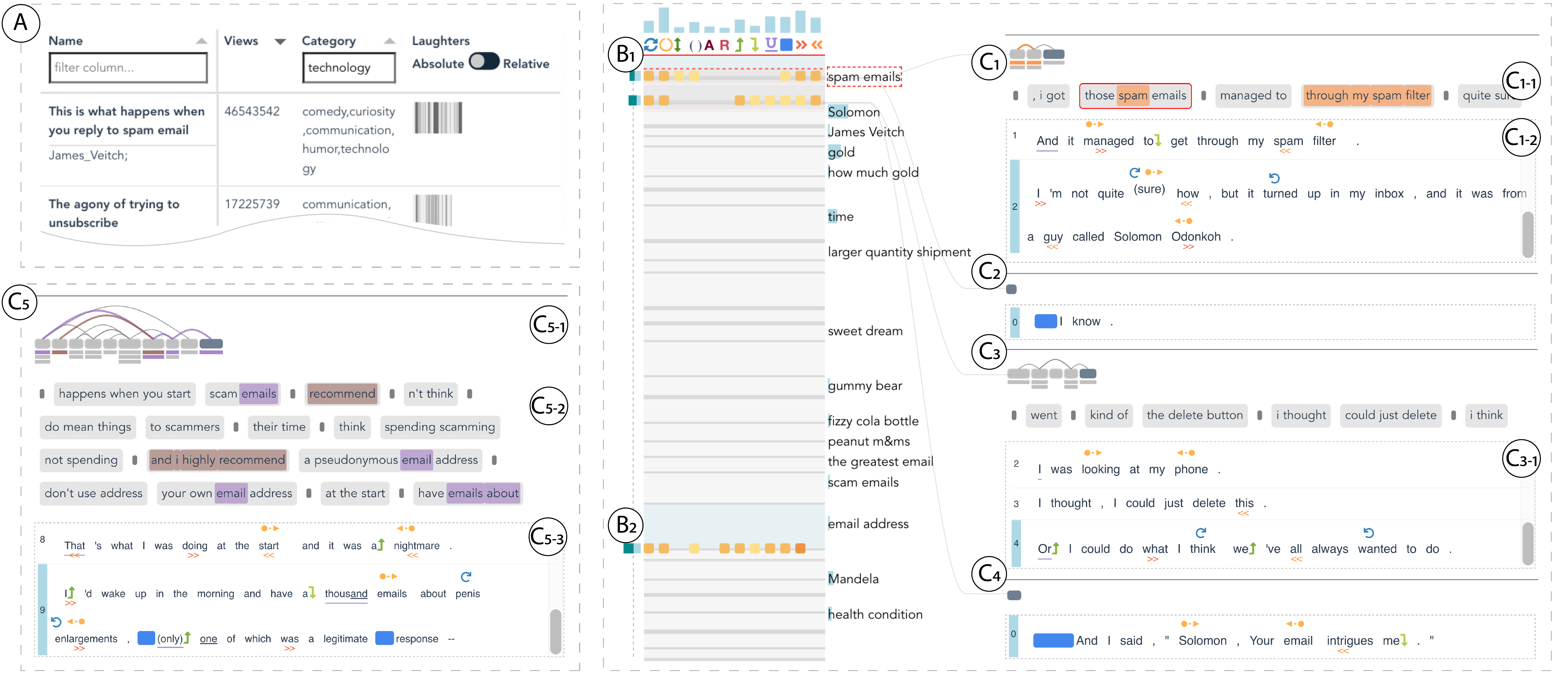}
\vspace{-3mm}
\caption{\xingbo{The case study on \textit{TED Talk}. After selecting the speech in (A), the user clicked on ``\textit{spam emails}'' in the augmented time matrix. Then the snippet in the speech opening (B1) and their context linking graphs (C1-4) are shown. Next, the user used the {\hf} (Fig.~\ref{fig:teaser}B) to find the snippet in the speech closing (B2) with rich humor delivery skills (long dark green bars to the left). Its context linking graph is shown in (C5).}}
\label{fig:ted_case}
\vspace{-4mm}
\end{figure*}

\subsection{Case Study on TED Talks}
\label{subsec:case2}
In the second case, the communication coach \textit{C1} explored humor skills in \textit{TED Talk} speeches that are related to ``technology''.
Since lots of his clients come from IT companies,
he expected talks on technologies are suitable teaching examples for using humor in speech.
In particular, he 
focused more on the timing of humor and speakers' vocal delivery skills, 
which were regarded as practical and effective
humor skills for students to follow and further improve their speeches.
Sorting the \xingbo{speeches} by the number of views in descending order, he discovered the most popular \xingbo{speech} named \emph{``This is what happens when you reply to spam email''\footnote{Ted Talk video url: \url{https://bit.ly/3eFqm6P}}}. 

\textbf{\textit{Case Context}}: 
This case includes two pieces of the speaker's experiences of replying to spam emails.
In the speech opening, the speaker introduced that he once received an email from a sender who had a strange name, and described how he replied to the email for fun.
To wrap up the speech, the speaker first suggested the audience replying to spam email with a pseudonymous email address. 

Originally, \textit{C1} noticed that in the bar code chart (Fig.~\ref{fig:ted_case}A), 
the laughter has a dense concentration at the start and end of the timeline, which was often considered as a pattern of strong opening and closing. He clicked the \xingbo{speech} to saw how the speaker delivered humor (\textit{\textbf{R1, R4}}) to entertain the audience at the start and the end (\textit{\textbf{R2}}).

\subsubsection{Speech opening}

He noticed that ``spam emails'' appears near the top of augmented time matrix (Fig.~\ref{fig:ted_case}B1). He clicked the phrase and saw how the speaker introduced it.
\casetag{Context relationship analysis \& Humor context}
From the highlighted phrases \example{those spam emails} and \example{through my spam filter} in the context linking graph (Fig.~\ref{fig:ted_case}C1-1) (\textit{\textbf{R2, R4}}), he inferred that the speaker received a spam email.
Then, he clicked on the highlighted phrases and confirmed his thought after reading the detailed sentences.
\casetag{Punchline}
In the punchline, \textit{E1} found that the speaker introduced the identity of the spam email sender, who had a very strange name---``Solomon Odonkoh'' (Sentence \#2 of Fig.~\ref{fig:ted_case}C1-2).
\casetag{Feature analysis}
Specifically, \textit{E1} observed that the speaker slowed down his speed on the word ``guy (\textcolor{slow}{\faAngleDoubleLeft})'' before revealing the spammer's name.
Moreover, the speaker paused (\textcolor{pause}{\faSquare}), and immediately added a short phrase ``I know'' (Sentence \#0 in Fig.~\ref{fig:ted_case}C2), which triggered another immediate laughter about the stranger for the second time.
\textit{C1} commented that the pause and speed variation motivated the audience to think about the spammer's name and identity. 

Similarly, \textit{C1} also spotted a pause (\textcolor{pause}{\faSquare}) between the next two snippets (Fig.~\ref{fig:ted_case}C3, C4) (\textit{\textbf{R1, R4}}), so he explored them accordingly. 
\casetag{Humor context}
He discovered that after introducing the spammer, the speaker also considered deleting the email (Sentence \#3 in Fig.~\ref{fig:ted_case}C3-1). However, he decided not to. Instead, he did what \example{we've always wanted to do} (Sentence \#4 in Fig.~\ref{fig:ted_case}C3-1)---reply to this email.
\casetag{Punchline}
Then, the speaker shared his funny responses to the email, starting with acknowledgment, \example{Solomon, you email intrigues me.} (Sentence \#0 in Fig.~\ref{fig:ted_case}C4).
\casetag{Feature analysis}
\textit{C1} commented that this was a smart pause (\textcolor{pause}{\faSquare}) at the beginning for helping
engage audiences to digest the speaker's previous sentence.
Here the audience got a chance to connect \example{we've all always wanted to do} (the bottom of Fig.~\ref{fig:ted_case}C3) with their desire for replying to spam emails.
The pause aroused the audience's interests in the speaker's next move, which
enhanced the humorous effect of the speaker's unexpected acknowledgment of the spam email
(Sentence \#0 in Fig.~\ref{fig:ted_case}C4).

\subsubsection{Speech closing}
Then, \textit{C1} wanted to see more snippets at the end, with rich delivery skills, especially with pauses. Thus, \textit{C1} used the {\hf} to
find the highlighted snippet (\textit{\textbf{R2}}) (Fig.~\ref{fig:ted_case}B2).
For the first one, he referred to its context linking graph (Fig.~\ref{fig:ted_case}C5).
\casetag{Humor context}
As he tracked the repetition links (Fig.~\ref{fig:ted_case}C5-1) from the left to the right, he realized that the speaker expressed a positive attitude towards replying to spam email through repeated phrases \example{(highly recommend)} (green rectangles in Fig.~\ref{fig:ted_case}C5-2) and their sentence contents. The speaker suggested using a
\example{pseudonymous email address} to do so and explained the reason in the punchline (Sentence \#9 in Fig.~\ref{fig:ted_case}C5-3). \casetag{Punchline} He once used his own email address. The result is that the mailbox was flooded with \example{a thousand} useless advertisements about \example{penis enlargements}. Among them, he was only able to find one legitimate information that he wanted (Sentence \#9 in Fig.~\ref{fig:ted_case}C5-3).
\casetag{Feature analysis}
The speaker stressed the reason in the punchline by pauses (\pause) and vocal stress (\stress) on keywords such as \example{thousand} and \example{only} (\textit{\textbf{R1, R4}}).

{\textit{C1}} commented that the speaker effectively used pauses to adapt the pace of his presentation to engage the audience, which is considered to be comic timing.
He added that pause is crucial to speech delivery, and most students do not realize how powerful it is. He emphasized that this speech is a good example for teaching students how to use pauses to deliver a strong opening and closing in their speeches.

\subsection{Expert Interviews}
\label{subsec:expert_interviews}

\xbRevise{We collected the experts' feedback from the individual interviews with the aforesaid experts (\textit{E1, E6, C1, and C3})}. Each interview lasted about one hour and was recorded with the participants' consent. First, we gave the participants a fifteen-minute tutorial outlining the humor features with concrete examples, as well as the visual designs and interactions of {\name}.
Then, participants were asked to explore the speeches introduced in
Secs.~\ref{subsec:case1} and \ref{subsec:case2} in a think-aloud manner for about forty minutes. For each speech, they \xbRev{were asked} to find and explore humor snippets with specific timing and features (\textit{\textbf{R2}})---snippets that contain (1) words-of-interest at speech opening and (2) humor-features-of-interest at speech closing. 
Then, within each snippet, they were required to reason about what contributes to the humor. Specifically, they were assigned the following tasks:

\begin{compactenum}
\item To examine if our context linking graph effectively highlights related build-ups (\textit{\textbf{R3}}), we asked participants to summarize how the speaker builds up humor.
\item To evaluate our inline feature highlighting (\textit{\textbf{R1, R4}}), we asked participants to identify which part of the punchline contributes the most to laughter in terms of word usage and vocal delivery.
\item To validate extracted features (\textit{\textbf{R4}}), we asked participants to read the original text script, listen to the audio, and voice any features that are out of place.
\end{compactenum}
\xbRevise{We then collected post-study feedback on system designs, usefulness, usability, and suggestions for improvements.}

\subsubsection{Results}

\xingbo{Compared with manual browsing and digesting of raw humor speeches, all the participants confirmed the usefulness of computation ability and visualization in the system for assisting humor exploration and analysis.

\xb{\textbf{Concrete humor representations}}.
The participants appreciated that {\name} automatically segments a speech based on the audience laughter. They confirmed it helps them quickly focus on the highlights of humor.
And the system offers convenient and user-friendly functions for revealing humor patterns in both speech content and vocal delivery.
The context linking graph was generally considered useful for traversing and summarizing humor build-ups.
\textit{E1} said, ``\textit{The context summary (at the top) helps understand and track the backbones of the story.}''
They praised 
the straightforward inline annotations of textual and audio features.
These annotations help the participants quickly identify the important word use, utterances, and their co-occurrence for creating humor; on the contrary, it is challenging to capture these patterns by watching videos only. 
\textit{E6} mentioned that ``\textit{these annotations, especially the audio feature annotations, successfully guide my attention (to critical parts of the punchline).
They vividly capture the delivery patterns within the sentence. I can picture the speaker in front of me giving a speech!}''

\textbf{Analysis flow}.
\xbRevise{For each speech, the participants explored around nine minutes of the speech content for humor analysis.
They confirmed that multi-level humor exploration supported by \name{} aligns well with their general analysis workflow.
The most time-consuming task is humor context analysis. But all of the participants could finish it in about three minutes with \name{}. The punchline analysis took them about one minute, and the extracted features were validated within a minute.
Finally, the participants could elaborate on what contributes to the humor
in a snippet regarding the verbal content and vocal delivery.

During the exploration, the textual incongruity features of punchlines were frequently used to identify the essence of humor.
The pitch and pause were found most useful for revealing the key delivery patterns.
Also, the participants often relied on the verb and noun phrase repetitions
to gain an overview of the humor story development in a snippet.}

\textbf{Usage scenarios}.
The participants valued the interactive exploration experience with {\name} and were eager to use it in the future.
Coaches \textit{C1} and \textit{C3} believed it would be an excellent teaching tool for coaches to show their students how to impress the audience with concrete examples (\eg, where to pause).
\textit{E1} confirmed that {\name} provides a corpus with various humor examples and
enables rich interactions, which facilitates a systematic study of humor.
}

\xingbo{Despite the positive feedback above, our participants also identified several limitations of {\name} and provided some suggestions for improvement.

\textbf{Reliability of feature extraction.}
\xbRevise{Our participants found that the extraction of textual features, especially inter-sentence repetitions and incongruity, contains more errors than the extraction of audio features. For example, they did not \xbRev{find a strong} semantic disconnection between \textit{``f**king''} and \textit{``questions''} (Fig.~\ref{fig:teaser}C5). 
However, they thought that the false positives of incongruity usually did not affect humor analysis very much, since they \xbRev{could} highlight some critical
content words in punchlines for digesting humor context.
In contrast, the errors of inter-sentence repetitions \xbRev{might negatively affect}
their exploration experience. For instance, 
the phrases (\textit{``i got''}, \textit{``managed to''}, \textit{``quite sure''}) in Fig.~\ref{fig:ted_case}C1 
were not regarded as repetitions, and their presence confused the participants. \textit{E6} thought showing meaningless inter-sentence repetitions is a little distracting. Thus,
\textit{E6} was a bit suspicious about
the effectiveness of the context summary in the context linking graph, and she tended to directly explore the sentence details.}

\textbf{Learning curve.}
\xbRevise{Though a fifteen-minute tutorial was provided,
the participants still needed our guidance to finish some required tasks at the very beginning (i.e., the first ten to fifteen minutes) of their exploration.
For example, the participants \xbRev{might} not remember all the visual encodings and interactions, and we \xbRev{further explained} them.
When we illustrated the system designs, 
the participants found the augmented time matrix was the most complex view.
But
after several trials,
they could successfully utilize it to find snippets of interest for further humor analysis.}
They claimed it is worth the effort to learn the system features and were willing to use {\name} in their future research or work.

In addition, they have provided us with valuable suggestions. For example, \textit{E1} recommended adding a sentence comparison function to examine the nuances of vocal delivery or word usage in different sentences.
\textit{C1} commented that besides texts and voice, visualizing gestures and facial expressions can enhance the analysis of humor techniques.}
\section{Discussion}
Here, we discuss the lessons learned and system generality.
We also identify several limitations that need further research in future work, including extending humor features, alleviating algorithm inaccuracy, enhancing system scalability, and enabling personalized humor explorations.

\textbf{Lessons learned}.
We learned two important lessons during our system design and evaluation.
\textit{1) Social context is important for humor understanding.}
During the evaluation, experts pointed out that interpreting humor requires external knowledge of social context. For example, understanding the humor in Sec.~\ref{sec:introduction} needs to know the ``are you a cop'' trope in American movies and TV, and the one in Sec.~\ref{subsubsec:context_analysis} relates to the WWII history.
\textit{2) Compact summary of multimodal features is helpful for multimedia analysis.}
Given the multimodality and heterogeneity of humor expression in speeches, 
we present inline annotations of verbal and vocal features along the text. Experts confirmed the annotations help gain quick insights into speech content and vocal delivery of humor, as well as the relation between them.
We believe that the integration of visual representation and multimedia data facilitates intuitive multimedia content understanding.


\textbf{System generality}.
While {\name} supports the analysis of speech content and vocal delivery of humor based on audience laughter markers, 
it can also be extended to evaluate public speaking skills based on other types of audience reaction (\eg, booing).
For example,
by highlighting the audio features of the speech sentence in a snippet that elicits booing or applause, we can further investigate effective voice modulation skills.
\xbRevise{In addition},
when there is no audience audio, the context linking graph can \xbRev{still} be used for text analysis. First, the text can be divided into snippets based on paragraphs or text segmentation algorithms. Then, the context linking graph can visualize contextual repetitions and narrative flows within a text snippet in various forms of literature (e.g., poetry and novel).

\textbf{Extending humor features}.
In this work, 
we derived a set of significant textual and audio humor features to analyze the speech content and its vocal delivery.
The proposed features can be enriched and further improved to enhance the understanding of humor.
First, as suggested by our experts,
it is interesting to explore how features from other modalities 
contribute to the delivery of humor. For example, facial landmarks~\cite{hasan2019ur, petridis2009joke, petridis2013mahnob}, and head movements~\cite{petridis2009joke} have been mentioned in previous research. How to incorporate features from these modalities in the analysis is a challenging while promising direction.
Second, the extraction and visualization of the repeated phrases for the humor build-up can be enhanced.
\xingbo{Currently, we focus on inter-sentence repetitions between punchlines for humor context analysis.
During the expert interviews, some participants discovered that some interesting repetitions appear across different snippets and incongruous concepts are distributed across different sentences. They wished to explore them.
Hence, we plan to extend the contextual repetition algorithm to extract semantically dissimilar phrases between sentences and to highlight repetitions in the whole speech.}
\xbRevise{Additionally}, there are other textual features for humor context analysis.
For example,
funny riddles are used by many comedians and public speakers to entertain and interact with the audience.
We plan to extend context-level features to facilitate
further study
of a humorous story.

\textbf{Alleviating inaccuracy of feature computation}.
Through case studies and expert interviews, we showed that the computation and visualization of humor features assisted users in reasoning about humor styles.
Inevitably, the imperfection of the algorithms will have harm the effectiveness of {\name}.
To alleviate such issues, we will keep improving the feature extraction.
Specifically, we plan to label humor features in the sentences 
and train advanced deep learning models (\eg,
BERT~\cite{devlin2018bert}, GPT-3~\cite{brown2020language}) for more accurate computation.
Moreover, we will improve the current visualization by encoding model uncertainty. For example, we can give visual hints (\eg, opacity) about the models' accuracy to alert users when the models output features with low confidence scores.

\textbf{Enhancing system scalability}.
\xingbo{Our system divides a speech into snippets based on the laughter occurrences. When the transcript is too long with too many punchlines, the exploration of humor snippets will be not so effective. To mitigate such an issue, we can consider merging neighboring humor snippets based on the semantic similarity and temporal proximity.}
\xingbo{Moreover, with the increasing number of repetitions within a humor snippet,} the context linking graph may have visual clutter of links.
More advanced interaction techniques are needed to address such issues (\eg, allowing users to interactively reduce and control the visibility of different groups of links). 

\textbf{Enabling personalized humor explorations.}
{\name} helps users narrow down to a video of interest according to the speech title, speaker, category, and laughter occurrences. In addition, it provides visual cues for users to find humor snippets based on textual and audio features.
As suggested by our communication coaches in expert interviews, supporting more complex user queries (\eg, styles of humor feature combinations) would enable a more personalized exploration of humorous content. 

\xbRevise{\textbf{Improving system evaluation.} The current evaluation is conducted with only four expert users. 
A long-term study with more domain experts can further validate the usability and effectiveness of \name{}, which is left as future work.}

\section{Conclusion}
In this work, we presented {\name}, a visual analytics system for exploring and analyzing humorous snippets in \xbRev{public speaking}. We first summarized humor-related features and design requirements based on literature review and user interviews. Then we developed a set of methods for presenting and decomposing multimodal features from a humorous speech.
Through case studies on stand-up comedy shows and TED Talks, as well as interviews with domain experts,
we demonstrated the usefulness and usability of {\name} in helping users explore and analyze speech content and vocal delivery of humor in speeches.

\xbRevise{In future work, we can improve the system usability by supporting humor query and humor style comparison. 
We plan to integrate more contextual features and features from other modalities (\eg, facial expressions) into the system. We can also apply deep learning models to improve the feature extraction accuracy. Furthermore, we will conduct a long-term study with more experts to further evaluate the system usability and its effectiveness for humor analysis.}

\ifCLASSOPTIONcompsoc
  \section*{Acknowledgments}
\else
  \section*{Acknowledgment}
\fi

We would like to thank our industry collaborator, Own The Room Asia Limited, for offering valuable resources. We also thank
our domain experts and the anonymous reviewers for their insightful comments.
This project is partially funded by a grant from ITF UICP (Project No. UIT/142).

\ifCLASSOPTIONcaptionsoff
  \newpage
\fi



%

\newpage
\bibliographystyle{IEEEtran}
\bibliography{reference.bib}


%

\vspace{-10mm}
\begin{IEEEbiography}[{\includegraphics[width=1.1in,height=1.5in,clip,keepaspectratio]{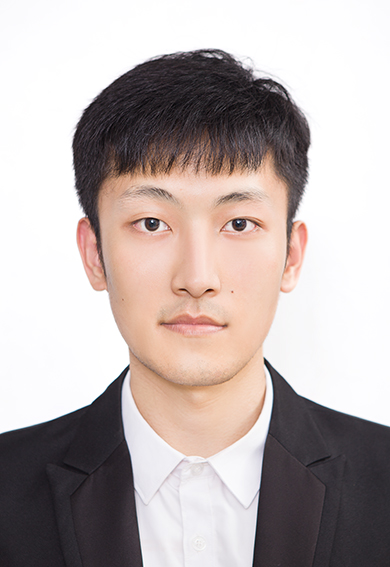}}]{Xingbo Wang}
is a Ph.D. candidate in the Department of Computer Science and Engineering at the Hong Kong University of Science and Technology (HKUST). He obtained a B.E. degree from Wuhan University, China in 2018.
His research interests include multimedia visualization, interactive machine learning for natural language processing (NLP).
For more details, please refer to \url{https://andy-xingbowang.com/}.
\end{IEEEbiography}

\vspace{-10mm}
\begin{IEEEbiography}[{\includegraphics[width=1.1in,height=1.5in,clip,keepaspectratio]{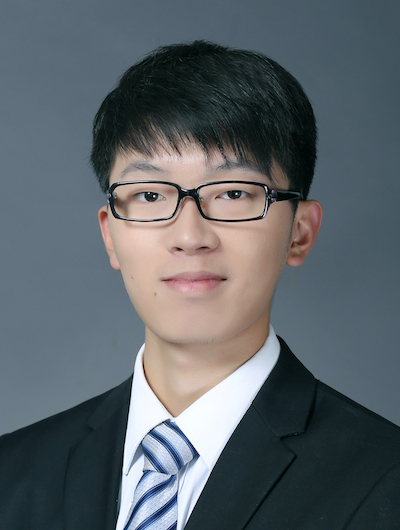}}]{Yao Ming}
is a research scientist at Bloomberg LP. His research focus on visual analytics, explainable machine learning, and natural language processing. He received a Ph.D. in Computer Science from the Hong Kong University of Science and Technology and a B.S. from Tsinghua University. For more details please refer to \url{https://www.myaooo.com}
\end{IEEEbiography}

\vspace{-10mm}
\begin{IEEEbiography}[{\includegraphics[width=1.1in,height=1.5in,clip,keepaspectratio]{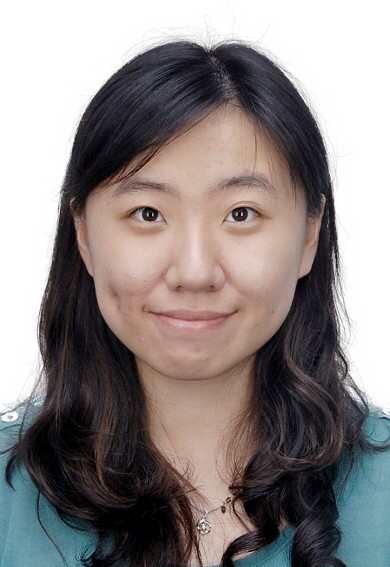}}]{Tongshuang Wu}
is a fifth year PhD student at the University of Washington, co-advised by Jeffrey Heer and Daniel S. Weld. 
She received her BE from the Hong Kong University of Science and Technology (HKUST).
Her research focuses on helping humans more effectively and systematically evaluate and interact with their models through Human-Computer Interaction (HCI) and Natural Language Processing (NLP).
For more details, please refer to \url{https://homes.cs.washington.edu/~wtshuang/}.

\vspace{-10mm}
\end{IEEEbiography}
\begin{IEEEbiography}[{\includegraphics[width=1.1in,height=1.5in,clip,keepaspectratio]{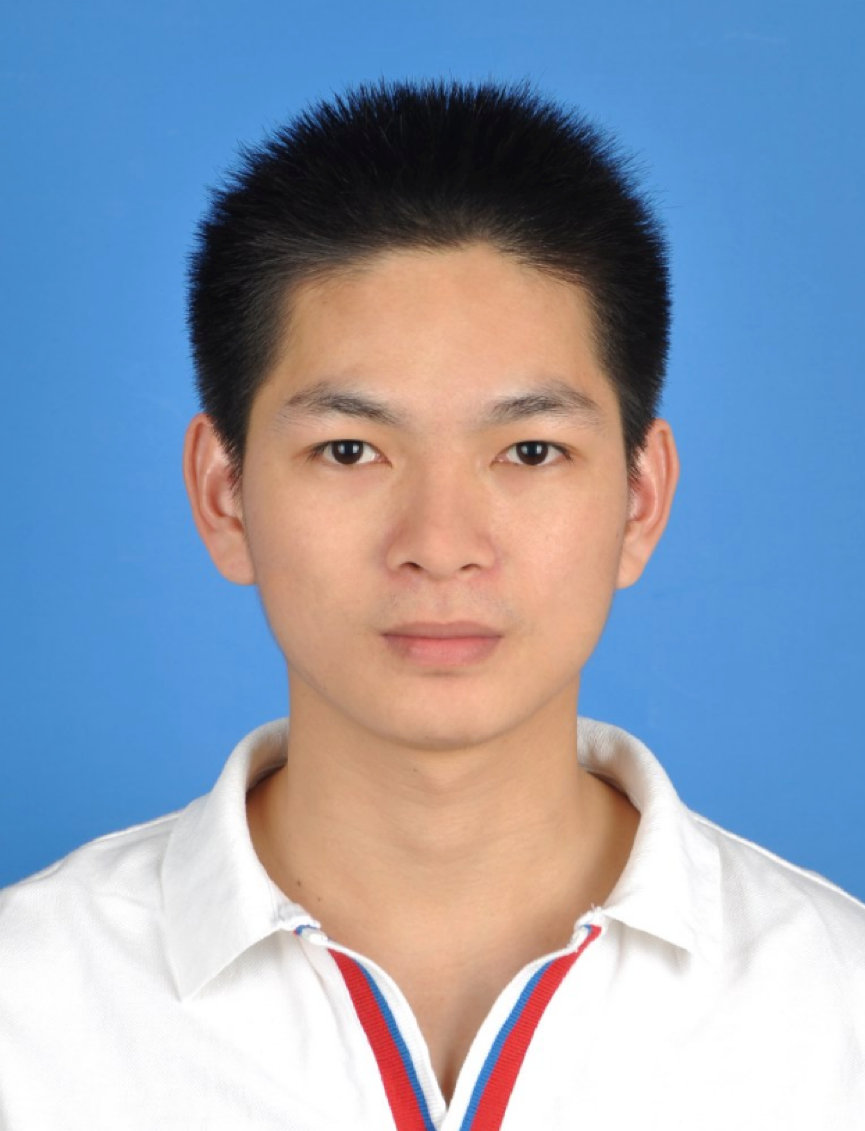}}]{Haipeng Zeng}
is currently an assistant professor in School of Intelligent Systems Engineering at the Sun Yat-sen University (SYSU). He obtained a B.S. in Mathematics from Sun Yat-Sen University and a Ph.D. in Computer Science from the Hong Kong University of Science and Technology. His research interests include data visualization, visual analytics, video analysis and machine learning.
\end{IEEEbiography}

\vspace{-10mm}
\begin{IEEEbiography}[{\includegraphics[width=1.1in,height=1.6in,clip,keepaspectratio]{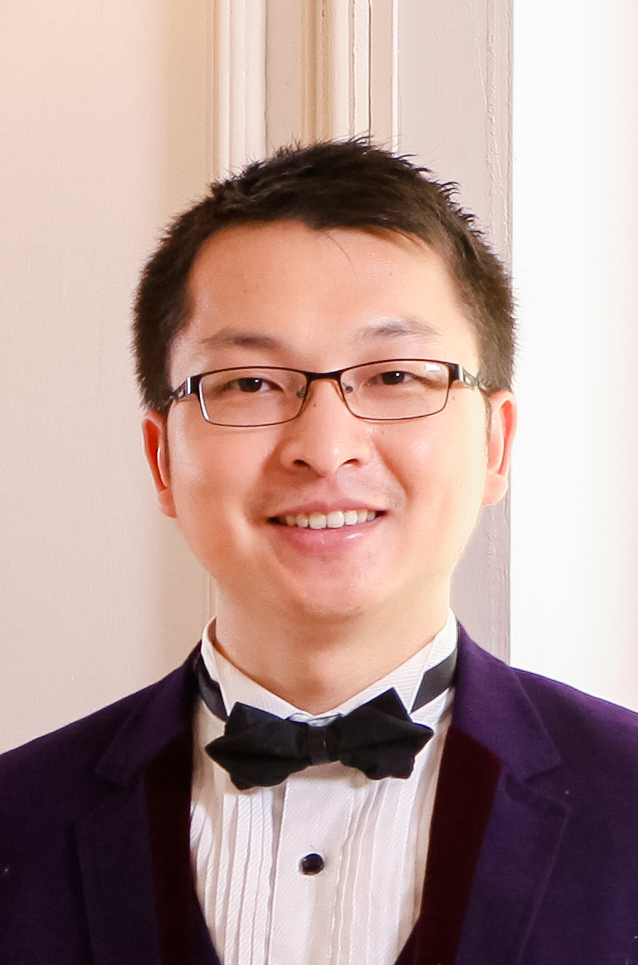}}]{Yong Wang}
is currently an assistant professor in School of Information Systems at Singapore Management University. His research interests include data visualization, visual analytics and explainable machine learning.
He obtained his Ph.D. in Computer Science
from Hong Kong University of Science and Technology in 2018. He received his B.E. and M.E. from Harbin Institute of Technology and Huazhong University of Science and Technology, respectively.
For more details, please refer to \url{http://yong-wang.org}.
\end{IEEEbiography}

\vspace{-10mm}
\begin{IEEEbiography}[{\includegraphics[width=1.1in,height=1.5in,clip,keepaspectratio]{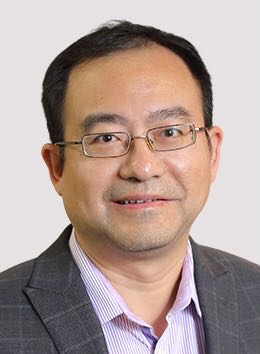}}]{Huamin Qu}
is a professor in the Department of Computer Science and Engineering (CSE) at the Hong Kong University of Science and Technology (HKUST) and also the director of the interdisciplinary program office (IPO) of HKUST. He obtained a BS in Mathematics from Xi'an Jiaotong University, China, an MS and a PhD in Computer Science from the Stony Brook University. His main research interests are in visualization and human-computer interaction, with focuses on urban informatics, social network analysis, E-learning, text visualization, and explainable artificial intelligence.
\end{IEEEbiography}



\end{document}